\documentclass[accepted, specialissue]{melba}
\usepackage{amsmath,amsfonts}

\usepackage{array}
\usepackage{pgfplots}
\pgfplotsset{compat=1.18}
\usepackage{float}

\usepackage[table]{xcolor}
\usepackage{multicol}
\usepackage{caption, subcaption}

\usepackage{balance}

\hypersetup{colorlinks, citecolor=[rgb]{0.0,0.7,0.5}, filecolor=[rgb]{0.6,0.0,0.0}, linkcolor=[rgb]{0.0,0.7,0.5}, urlcolor=[rgb]{0.0,0.0,0.5}}

\melbaid{2026:031}  % This is provided upon by the publishing editor
\doi{10.59275/j.melba.2026-cbeb}
\melbaauthors{Ward and Imran}  % Note: this one is also used to set the pdf 'authors' metadata
\email{aimran@uky.edu}
\volume{2026}
\firstpageno{634}  % Communicated by the publishing editor
\melbayear{2026}  % The publication year
\datesubmitted{2025-12-29}  % Date submitted to MELBA: mm/yyyy
\datepublished{2026-09-19}  % Today's date: mm/yyyy

\melbaspecialissue{Medical Imaging with Deep Learning (MIDL) 2025}
\melbaspecialissueeditors{Lisa Koch, Ronald M. Summers, Chen Chen, Yan Zhuang}

\ShortHeadings{ABFR-KAN: Kolmogorov-Arnold Networks for Functional Brain Analysis}{Ward and Imran}

\title{ABFR-KAN: Kolmogorov-Arnold Networks for Functional Brain \\Analysis}

\author{
	\firstname Tyler \surname Ward\orcid{0000-0003-0669-1407},
	\firstname Abdullah \surname Imran\orcid{0000-0001-5215-339X}
}
\affiliations{% <- trailing '%' to avoid unwanted indent
	\addr Department of Computer Science, University of Kentucky, Lexington, KY 40506, USA
}

\abstract{Functional connectivity (FC) analysis is widely used for computer-aided brain disorder diagnosis, but conventional atlas-based approaches can introduce structural bias and overlook subject-specific brain function. Addressing this, we propose \textbf{A}dvanced \textbf{B}rain \textbf{F}unction \textbf{R}epresentation with \textbf{K}olmogorov-\textbf{A}rnold \textbf{N}etworks (\textbf{ABFR-KAN}), a transformer-based framework that combines globally shared randomized anchor patches with subject-specific multi-scale patch sampling and Kolmogorov–Arnold Networks (KANs). The randomized anchors reduce structural bias and improve anatomical conformity while providing a consistent reference space across subjects. In contrast, per-subject patch sampling captures individual functional variability and improves the reliability of FC estimation. We evaluate the proposed method on resting-state functional magnetic resonance imaging (rs-fMRI) data from the Autism Brain Imaging Data Exchange (ABIDE) I and II datasets (281 subjects and 211 subjects, respectively). Extensive experiments, including 5-fold cross-validation, cross-site evaluation, and ablation studies across varying model backbones and KAN configurations, demonstrate that ABFR-KAN consistently outperforms state-of-the-art baselines for autism spectrum disorder (ASD) classification. These results highlight the potential of combining advanced brain function representation strategies with KAN-based architectures for robust neuroimaging analysis. Our code is available at \url{https://github.com/tbwa233/ABFR-KAN}.}


\begin{document}

\twocolumn[\maketitle]

\section{Introduction}
\enluminure{A}{CCURATE} and early diagnosis of brain disorders is crucial for effective medical intervention and treatment planning \cite{kumar2025early}. A widely adopted method for aiding diagnosis of such conditions is the analysis of functional connectivity (FC), which measures relationships between brain regions using blood-oxygen-level-dependent (BOLD) signals captured during resting-state functional magnetic resonance imaging (rs-fMRI) \cite{du2018classification}. However, FC analysis (FCA) often relies on pre-defined atlases to define regions-of-interest (ROIs) within the brain, which carry several limitations. This approach can lead to subjective selection bias, disregard for individual specificity, and a lack of interaction between brain regions and FCA \cite{liu2024randomizing}. Despite research into various methods of addressing these issues, such as data-driven \cite{reeves2025fmri}, individualized \cite{hakonen2025individual}, and multi-atlas \cite{han2025dual} setups, a definitive resolution to all of the challenges associated with atlas-based parcellation techniques has not yet emerged.

Additional drawbacks of traditional FCA methods are the high dimensionality and complexity of the functional representations, and finding solutions addressing these issues is an active area of research \cite{wang2018neurophysiological, vidaurre2021a}. In the age of artificial intelligence (AI), many researchers have explored FCA methods that incorporate deep learning (DL) \cite{qu2021brain}, such as multi-layer perceptrons (MLPs) that take a flattened FC matrix as an input \cite{hou2025brainnetmlp}. MLP-based methods have demonstrated effectiveness in downstream tasks such as brain disorder classification \cite{sachdeva2024resolving}, but they share similar drawbacks as non-AI-based methods for FCA.

For example, a connectome for $N$ brain regions is often represented by an $N\times N$ matrix or an $\frac{N(N-1)}{2}$-length feature vector for unique pairs \cite{kaiser2011tutorial}. This means that for a moderately sized parcellation (e.g., $N = 100$), there could be upwards of 10,000 features per subject. This makes MLPs trained on data such as this highly prone to overfitting due to the curse of dimensionality, as the extremely high-dimensional feature space requires estimating a vast number of parameters from relatively limited samples \cite{smucny2022deep}. In addition to this risk of overfitting, it is computationally burdensome to use MLPs for FCA, as MLPs that take a full FC matrix as input must process all pairwise connections simultaneously, which is inefficient. 

Recently, Kolmogorov-Arnold Networks (KANs) \cite{liu2024kan} have emerged as an alternative to traditional MLPs, leveraging learnable activation functions on edges rather than fixed activation functions on nodes. Inspired by the Kolmogorov-Arnold representation theorem, KANs replace conventional weight matrices with univariate functions parameterized as splines, offering improved expressiveness and flexibility in function approximation \cite{liu2024kan}. This design enables KANs to model complex transformations more efficiently while maintaining better interpretability and scaling properties compared to  \cite{liu2024kan}. Additionally, KANs have demonstrated potential in computer vision-related tasks \cite{cheon2024demonstrating}. Based on this, we hypothesize that replacing MLPs with KANs in brain disorder diagnosis models can better capture intricate relationships in FC patterns, leading to more robust and individualized diagnoses.

\begin{figure}[t]
    \centering
    \includegraphics[width=\linewidth]{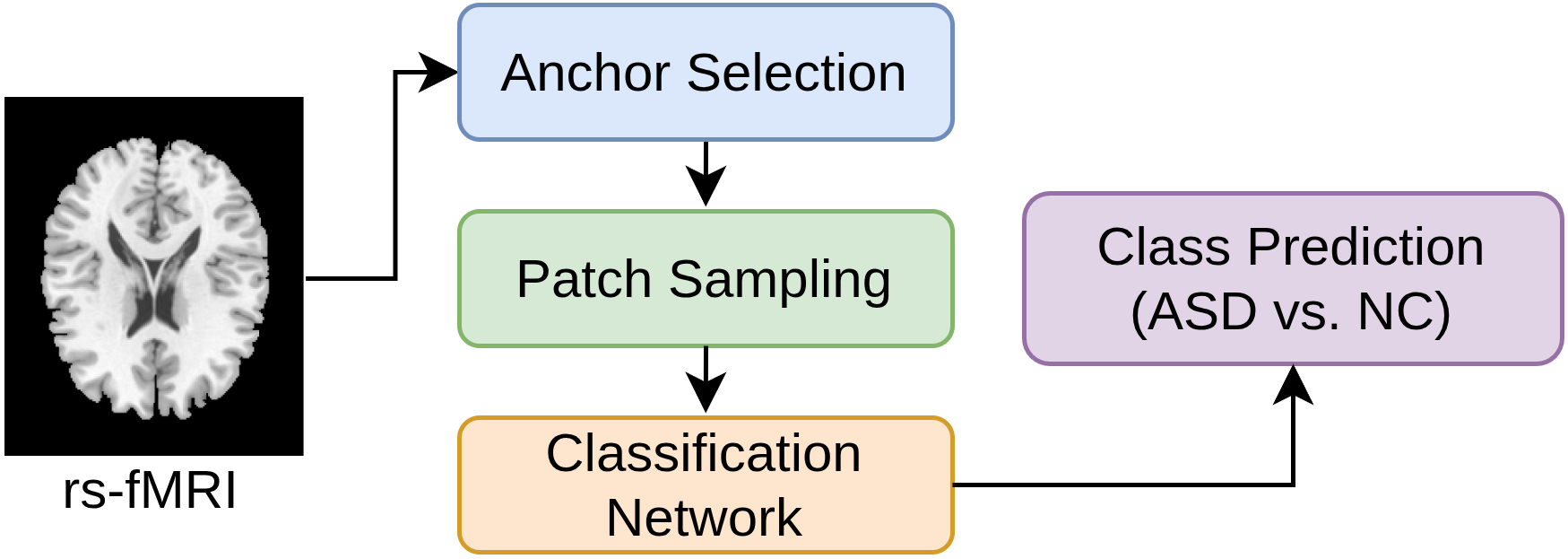}
    \caption{The general workflow of our proposed ABFR-KAN framework. In the anchor selection stage, a fixed set of reference patches is randomly sampled from grey matter (GM) regions to serve as anatomically unbiased functional anchors. In the patch sampling stage, multiple sets of brain patches are sampled at different spatial scales from each subject, and their FC to the anchors is computed to form robust, position-aware functional representations. These representations are then embedded with spatial information and processed by the classification network to produce a binary prediction of whether a subject belongs to the autism spectrum disorder (ASD) or normal control (NC) cohort.}
    \label{fig:teaser}
\end{figure}

In this paper, we propose \textit{ABFR-KAN} (\textit{A}dvanced \textit{B}rain \textit{F}unction \textit{R}epresentation with \textit{K}olmogorov-\textit{A}rnold \textit{N}etworks), a novel workflow for brain disorder diagnosis (Fig.~\ref{fig:teaser}). Building upon state-of-the-art methods, we propose novel sampling and function representation strategies and investigate the impact of KANs under various configurations in transformer networks. Our specific contributions in the present paper are summarized as:
\begin{itemize}
  \item Randomized anchor patch selection, which helps avoid structural bias, boosts individual-specific representations, and increases robustness and generalizability by reducing dependence on atlas-based parcellation.
  \item Multi-scale sampling of patches from a subject's brain, aimed at creating multiple functional representations for the same subject, introducing variance while preserving meaningful FC information. 
  \item Extensive experimentation demonstrating the effectiveness of replacing traditional MLP components in transformer networks with KANs.
\end{itemize} 
This manuscript is an extension of our paper ``Improving brain disorder diagnosis with advanced brain function representation and Kolmogorov-Arnold Networks'' presented at the 2025 Medical Imaging with Deep Learning (MIDL) conference~\cite{ward2025improving}. This manuscript substantially extends the work by adding a thorough literature review, additional experiments across new model backbones and KAN variants, and a more detailed description of the methods and results discussion.

%%%%%%%%%%%%%%%%%%%%%%%%%%%%%%%%%%%%%%%%%%%%%%%%%%%%%%%%%%%%%%%%%%%%%%%%%%%
% Related works
%%%%%%%%%%%%%%%%%%%%%%%%%%%%%%%%%%%%%%%%%%%%%%%%%%%%%%%%%%%%%%%%%%%%%%%%%%%
% Make sure to put your work into context and include apporpriate citations.
% We do not have limits on citation counts.
\section{Related Works}
\subsection{Brain Function Representation}
There generally exist three different setups for brain disorder analysis using an atlas: single-atlas, multi-atlas, and individual-specific atlas. In the single-atlas setup, a fixed, predefined brain atlas is used to partition the brain into ROIs, and FC is computed between these regions. This results in a consistent graph structure across subjects, enabling straightforward comparison but potentially limiting representational flexibility due to reliance on a single spatial scale and fixed anatomical definitions \cite{dickie2017whole}. An example of a single-atlas approach is BrainGNN \cite{li2021braingnn}, which constructs subject-specific brain graphs using the Desikan-Killiany \cite{desikan2006automated} atlas, where nodes correspond to ROIs and edges represent pairwise functional connectivity derived from fMRI time series. The model then applies graph neural network (GNN) operations along with ROI-aware convolution and pooling mechanisms to learn discriminative biomarkers while maintaining interpretability.

In contrast, the multi-atlas setup leverages multiple atlases simultaneously to capture complementary views of brain organization at different spatial resolutions or anatomical perspectives. Since no single atlas fully characterizes the complexity of brain structure and function, combining multiple atlases can provide richer and more robust representations \cite{dickie2017whole}. In this setting, FC networks are constructed separately for each atlas and then fused at either the feature level or the graph level to form a more comprehensive representation. An example of this can be seen in the work of \cite{lee2024spectral}, who leveraged a spectral GNN-based framework that constructs a holistic functional connectivity network by integrating ROI signals across multiple atlases \cite{kennedy1998gyri, craddock2012whole, rolls2020automated}, capturing both intra-atlas relationships and inter-atlas interactions. Their approach incorporates both early fusion (combining ROI signals before graph construction) and late fusion (aggregating model outputs), enabling the model to exploit multi-scale connectivity patterns for improved disease classification.

The individual-specific atlas setup aims to overcome limitations of group-level atlases by tailoring brain parcellations to each subject. Instead of relying on fixed ROI definitions, this approach constructs personalized functional regions based on subject-specific fMRI data, thereby capturing inter-subject variability in brain organization. This can lead to more precise and discriminative functional connectivity representations, particularly for neurological disorders where individual differences are significant \cite{taimouri2025characterizing}. For example, PFC-DBGNN-STAA by \cite{cui2023personalized} first generates a personalized FC template for each subject by aligning ROIs based on their individual FC patterns. It then combines both individual-specific and group-level representations within a dual-branch GNN and further enhances them using spatio-temporal attention mechanisms to capture both spatial dependencies and temporal dynamics in fMRI signals.

As an alternative to using pre-defined atlases for ROI parcellation, several data-driven approaches have been proposed. For example, attention-guided hybrid deep learning networks have been used to localize discriminative brain regions automatically for Alzheimer's disease and MCI diagnosis \cite{lian2020attention}. RandomFR \cite{liu2024randomizing} is an innovative approach for brain function representation and operates via a randomized selection of brain patches as well as the use of novel function and position description methods. RandomFR serves as the main inspiration for the research presented in this paper.

\subsection{Kolmogorov-Arnold Networks}
For decades, there has been debate regarding the optimal number of hidden layers/nodes when designing neural networks (NNs) \cite{stathakis2009many}. Such open questions regarding hidden layers have contributed to the lack of interpretability within NNs, thus giving them their reputation of being \textit{``black boxes"} \cite{tan2015improving}. The Kolmogorov-Arnold representation theorem has long been thought to explain the use of more than one hidden layer in NNs \cite{hecht1987kolmogorov}, although this is not a universally held belief \cite{schmidt2021kolmogorov}, and many investigations of building NNs based on this theorem occurred before the rise of DL \cite{liu2024kan}. Addressing this, \cite{liu2024kan} introduced their aptly named Kolmogorov-Arnold Network, or KAN, which provided a generalized form of the original Kolmogorov-Arnold representation to arbitrary widths and depths, boosting its practicality in a world driven by DL.

The core idea of the KAN framework is fundamentally different from traditional MLPs. In an MLP, learnable linear weight matrices are applied along edges, followed by fixed nonlinear activation functions at nodes. In contrast, KANs eliminate linear weights entirely. Each connection is instead parameterized by a learnable univariate function, and nodes perform only summation of incoming signals. Thus, KANs are not simply MLPs with learnable activation functions, but rather replace all scalar weights with function-valued parameters defined on edges. In practice, each edge encodes a univariate function implemented as a B-spline with trainable coefficients, and a KAN layer can be viewed as a matrix of such functions. The network is then formed as a composition of these nonlinear function matrices.

Because KANs learn both the compositional structure of multivariate functions via network connectivity and flexible univariate transformations via spline-parameterized edge functions, they achieve high expressiveness. At the same time, each learned component corresponds to an explicit, low-dimensional function that can be directly visualized or approximated symbolically, improving interpretability compared to standard MLPs. However, this expressive design introduces computational inefficiencies. Unlike MLPs, where each connection involves a single scalar multiplication, KANs require evaluating a learnable function on every edge. As a result, intermediate representations must store and propagate the outputs of many per-edge function evaluations rather than simple weighted sums, increasing both memory usage and computational cost during both forward and backward passes.

Addressing these performance bottlenecks, many improvements of KANs have been proposed. Particularly, in terms of identifying more effective and efficient activation functions compared to the initial residual activation function. One such improvement lies in the Efficient-KAN \cite{blealtan2024efficient} architecture, which leverages the linearity of B-spline basis functions instead of computing each spline activation independently for every edge via expensive tensor expansion. Specifically, Efficient-KAN restructures the computation by applying the fixed basis functions to the input once and then combining the results via a learnable weight matrix, akin to matrix multiplication. This avoids high-dimensional tensor expansions and makes the forward and backward passes both memory- and compute-efficient.

An alternative implementation, FastKAN \cite{li2024kolmogorov}, replaces the 3-order B-spline basis in the original KAN with radial basis functions (RBFs). \citet{li2024kolmogorov} demonstrates that this change has a positive impact, improving forward speed compared to Efficient-KAN by 33$\times$. This indicates that Gaussian RBFs are capable of approximating the B-spline basis, which is the bottleneck of KAN and Efficient-KAN.

FasterKAN, by \cite{delis2024fasterkan}, further improves the speed of KAN by using approximations of the B-spline via the reflectional switch function, inspired by \cite{deng2009conceptual}, modified to have reflectional symmetry. Designing the activation function in this manner allows FasterKAN to approximate the 3rd-order B-splines used in the original KAN implementation. Experimentation with this setup revealed that FasterKAN achieved speeds 1.5$\times$ faster than FastKAN. At the time this research was performed, FasterKAN was the most efficient KAN framework.

Since then, other KAN frameworks such as RBF-KAN \cite{sidharth2024rbf}, which uses RBFs as the basis function similar to FastKAN, and ChebyKAN \cite{guo2024chebykan}, which employs Chebyshev polynomials in place of B-splines, have demonstrated competitive performance with FasterKAN. Additionally, hybrid structures such as those combining B-splines with Gaussian RBFs \cite{ta2024bsrbf} or wavelet functions \cite{seydi2024unveiling} are interesting and have demonstrated good performance across a variety of tasks. Finding the best basis function and architecture for KAN is still an active research topic, and best practices are expected to remain dynamic and ever-changing.

Up to this point, our literature survey of KANs has focused on the history and development process. To shift focus to the application side, the original use case of KANs reported in \cite{liu2024kan} lay in improving AI + Science tasks such as function fitting, partial differential equation (PDE) solving, symbolic regression, and scientific discovery. The original authors, as well as subsequent work \cite{liu2024kan20}, have demonstrated the effectiveness of KANs in this regard, but it is far from the only use case.

Early work in expanding the applicability of KANs came in the form of assessing their suitability for computer vision tasks \cite{azam2024suitability}. It was discovered that on simple classification/segmentation tasks using the MNIST, CIFAR-10, and CamVid datasets, KAN-based vision models outperformed MLP and convolutional neural network (CNN)-based approaches in both accuracy and scalability, demonstrating their suitability for computer vision tasks. For computer vision applications, KANs have also demonstrated strong performance in medical imaging tasks \cite{li2025u}. Our work in this paper leverages VisionKAN \cite{chen2024vision}, a repository investigating whether KANs can replace MLPs in Vision Transformers. The next subsection will discuss the use of KANs specifically for imaging-based brain diagnostic tasks.

%%%%%%%%%%%
\begin{figure}[t]
    \centering
    \resizebox{\linewidth}{!}
    {
    \begin{tabular}{m{0.3\linewidth} m{0.3\linewidth} m{0.3\linewidth}}
    \includegraphics[width=\linewidth]{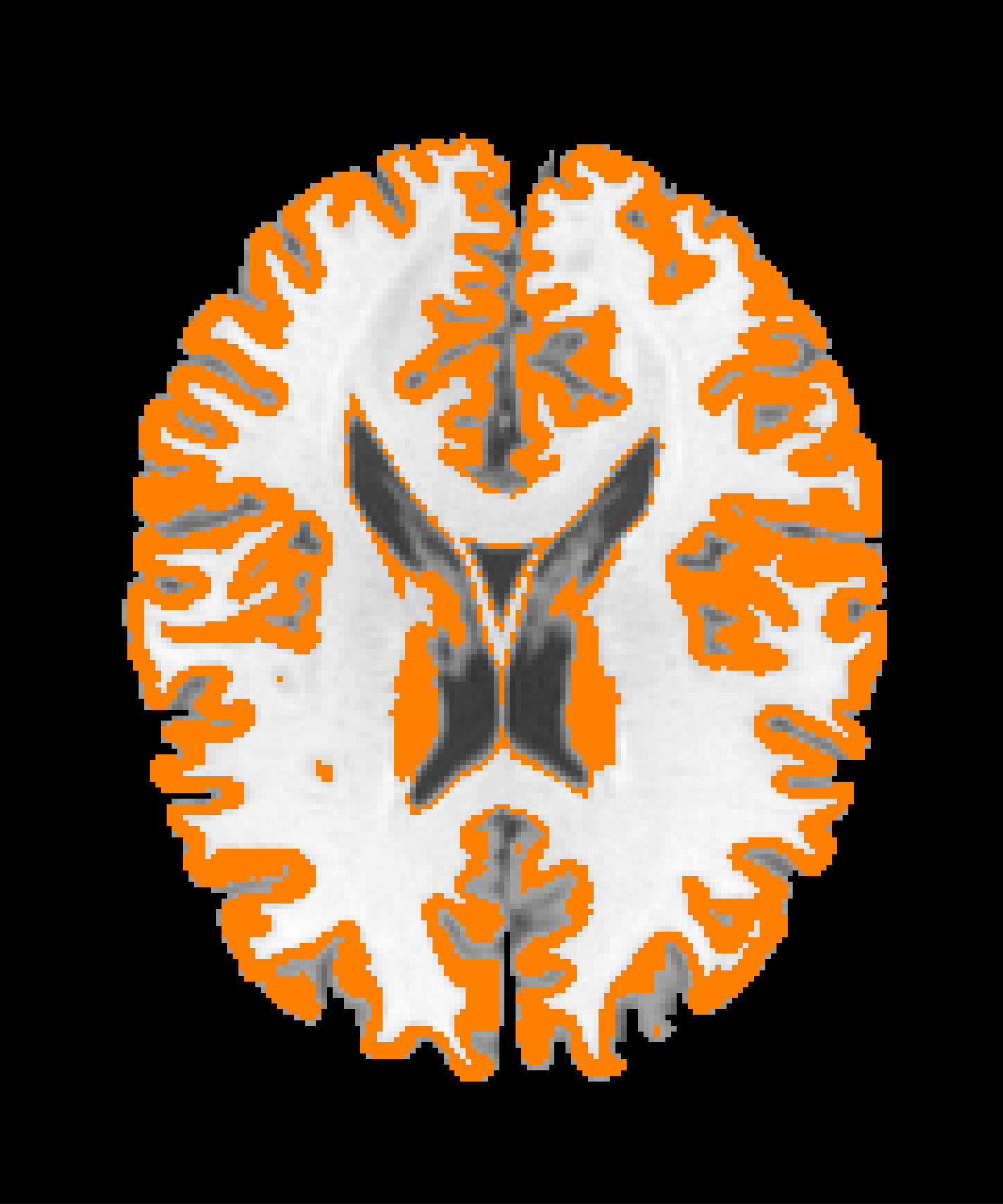} &
    \includegraphics[width=\linewidth]{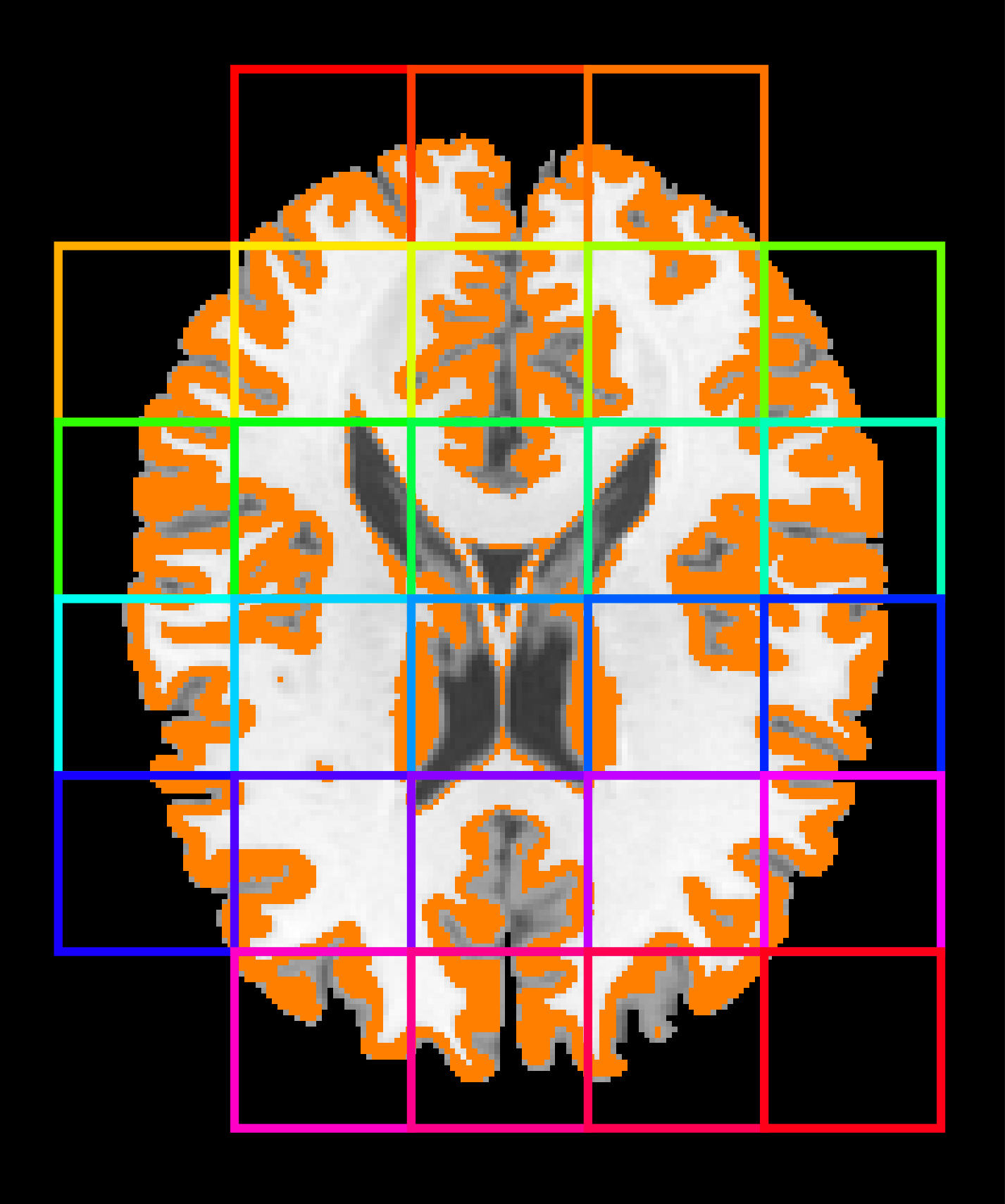} &
    \includegraphics[width=\linewidth]{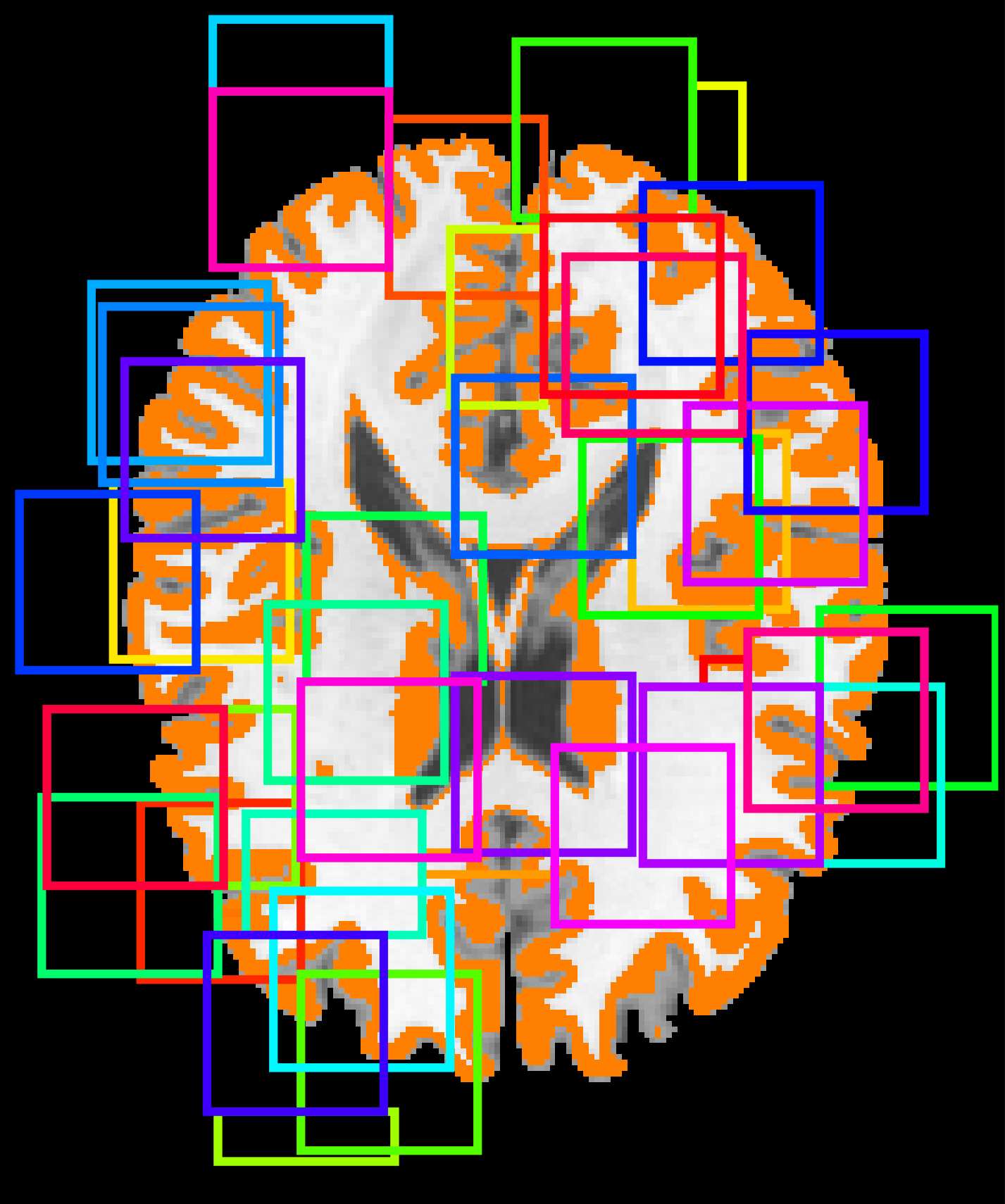}
    \\
    \centering\large\shortstack{(a) grey-matter \\ mask} &
    \centering\large\shortstack{(b) grid-based \\ anchor selection} &
    \centering\large\shortstack{(c) random \\ anchor selection}
    \end{tabular}
    }
    \caption{(a) The GM mask (highlighted in orange and overlaid over the \textit{ch2bet} template produced by \cite{holmes1998enhancement}) from which our anchor patches are selected. (b) The baseline grid-based anchor selection process. (c) Our randomized anchor selection process, which reduces structural bias and enhances individual specificity.}
    \label{fig:anchorselection}
\end{figure}
%%%%%%%%%%%%%%

\subsection{KANs for Brain Diagnostic Tasks}
\cite{wang2024pilot} explored the effectiveness of various KAN-based neural networks (KAN, FastKAN, and ChebyKAN) in detecting cognitive load from multi-channel EEG signals. They found that on a 19-channel EEG dataset constructed during a mental arithmetic task, the best KAN-based model outperformed the accuracy of existing state-of-the-art deep learning approaches like EEGNet and CNN-BiLSTM by 1.04\% and 3.22\%, respectively, with \citet{liu2024kan}'s original KAN implementation outperforming the FastKAN and ChebyKAN approaches. Additional KAN variants were explored on EEG data by \cite{thant2025emotion}, and it was found that Wavelet KAN outperformed the other KAN variants. \cite{hasan2024handeegnet} also explored the use of KANs on EEG data, designing a KAN-based model for hand movement classification from motor imagery EEG. The authors found that their KAN-based approach outperformed the conventional MLP-based approach by a magnitude of 19.5\%. Contrary to these studies, where the use of KAN-based architectures was found to be beneficial, \cite{hasan2025long} compared the performance of KAN to that of LSTMs for epileptic seizure prediction from EEG data and found the performance of KAN to be drastically below that of LSTM, by a margin of 15.22\% accuracy.

%%%%%%%%%%%
\begin{figure*}[t]
    \centering
    \subcaptionbox{ Grid Anchors vs. GM Surface}{\includegraphics[width=0.49\linewidth]{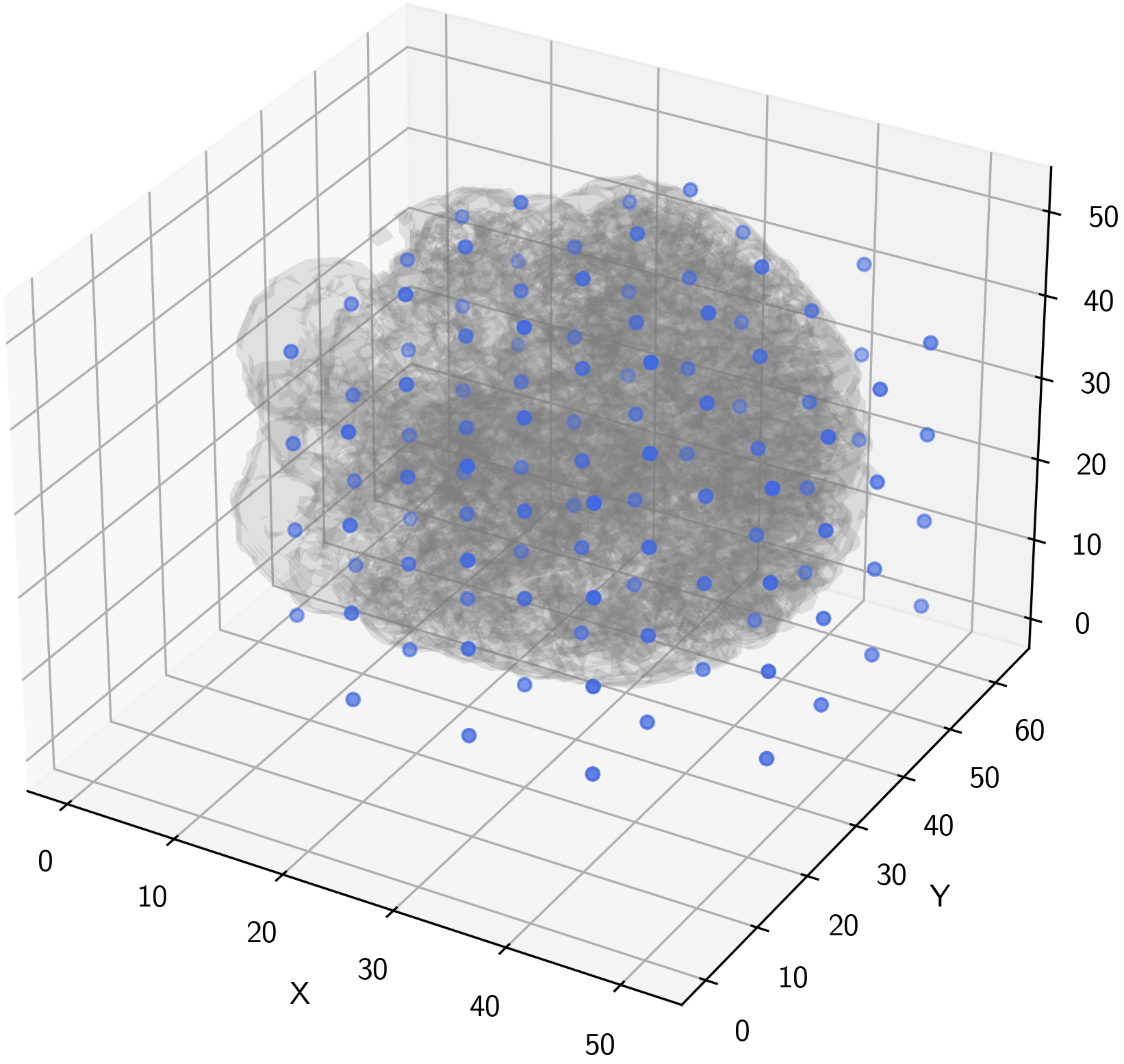}}
    \subcaptionbox{Random Anchors vs. GM Surface}{\includegraphics[width=0.49\linewidth]{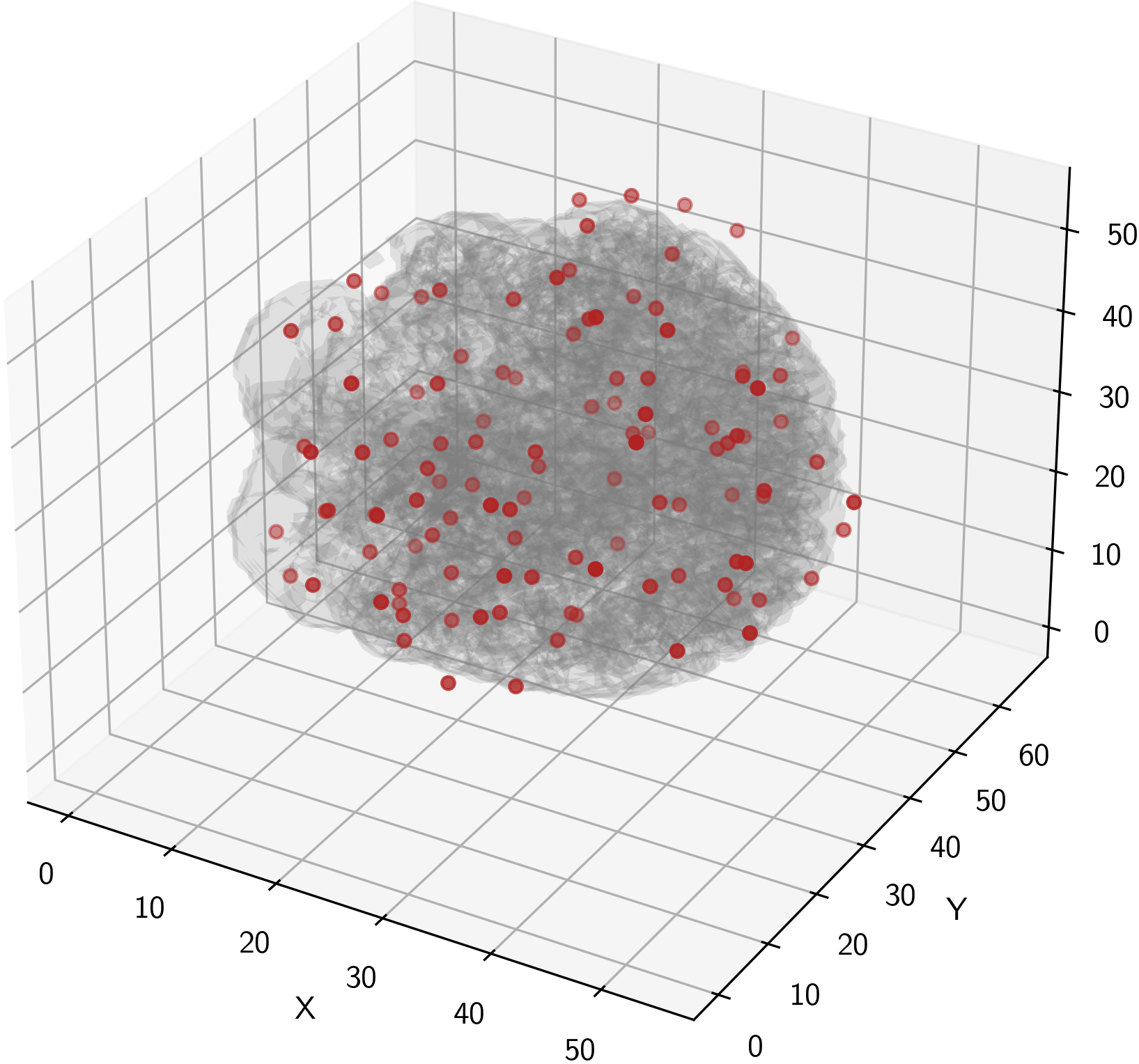}}
    
    \caption{Visualization of anchor patch centers overlaid on the GM mask (represented as a 3D point cloud of voxels). Anchors selected with a grid-based strategy (left) exhibit a rigid lattice distribution that extends into non-GM regions and fails to conform to cortical geometry. In contrast, randomly selected anchors (right) are distributed more adaptively within the GM and better follow the brain's anatomical shape.}
    \label{fig:anchorpatchcenters}
\end{figure*}
%%%%%%%%%%%%%

\citet{ding2025explainable} explored the efficacy of KANs for Alzheimer's disease (AD) diagnosis by integrating KAN into graph convolutional networks (GCNs) to enhance diagnostic accuracy and interpretability. Evaluated on the ADNI dataset, the authors found that their GCN-KAN model outperformed traditional GCNs by 4-8\% in classification accuracy while providing interpretable insights into key brain regions associated with AD. The applicability of hybrid KAN + GCN approaches for AD diagnosis was further validated by \cite{wang2025application}, who found that a KAN + GCN model outperformed traditional random forest (RF), support vector machine (SVM), MLP, and CNN approaches on the ADNI dataset. \citet{vermavgg} also explored KANs for AD diagnosis, integrating KAN layers as densely connected layers into the VGG-19 architecture. Their hybrid architecture, dubbed VGG-KAN, achieved a classification accuracy of 95.67\% on the OASIS dataset and an accuracy of 97.86\% on the ADNI dataset.

\citet{wang2025cest} demonstrated the feasibility of KAN and its variants for chemical exchange saturation transfer (CEST)-MRI data analysis. On a CEST-MRI dataset of 27 subjects, the authors explored the performance of MLP, KAN, and Lorentzian-KAN (LKAN) against a traditional multi-pool Lorentzian fitting (MPLF) method in generating various CEST parameters. They found that KAN and LKAN both outperformed MLP across the experiments. Sticking with MRI as the modality, \citet{penkin2024kolmogorov} examined the capabilities of three different KAN bottlenecks (linear, B-spline, and Chebyshev polynomials) as deep feature extractors for MRI reconstruction, finding that Chebyshev polynomials performed the best. \citet{giupponi2025accurate} evaluated the performance of KANs against CNNs for brain age prediction from MRIs, and found that KAN-based models reduced estimation errors, highlighting their potential for improving brain age assessment. Overall, these works span a wide range of tasks and model architectures, illustrating that KANs are not tied to a single network design but can be integrated into diverse neural architectures.

\section{Methods}

Following~\cite{liu2024randomizing} for brain function representation, our proposed ABFR-KAN workflow is divided into three stages: \textbf{sampling, function representation, and classification network}. In the sampling stage, anchor patches are selected from the GM region of rs-fMRI scans. The GM mask is obtained by segmenting the structural brain image aligned with the functional brain image. In the function representation stage, each sampled brain patch is described using two types of information: where it is located in the brain and how it functionally relates to other regions. The spatial location is encoded using its coordinates in Montreal Neurological Institute (MNI) space. To capture functional relationships, we compare the BOLD signal of the sampled patch with the BOLD signals of a set of anchor patches. Specifically, we compute the Pearson correlation between the time series of the sampled patch and each anchor patch. These correlation values form a vector that summarizes how strongly the sampled patch is functionally connected to all anchors. Collecting these vectors for all sampled patches produces a functional representation matrix, where each row corresponds to one sampled patch and each column corresponds to one anchor patch. In the classification network, embeddings based on the fusion of the function and position descriptions are passed to the classifier for final predictions.

\subsection{Random Anchor Selection}

\begin{figure}[t]
\resizebox{\linewidth}{!}{
\begin{tikzpicture}
\begin{axis}[
    ybar interval,
    bar width=0.8cm,
    width=11cm,
    height=7cm,
    xlabel={Distance to GM boundary (voxels)},
    ylabel={Count},
    ymin=0,
    ymax=45,
    xmin=0,
    xmax=8.5,
    xtick={0,1,2,3,4,5,6,7,8},
    ymajorgrids=false,
    grid style=dashed,
    axis x line*=bottom,
    axis y line*=left,
    xtick pos=left,
    ytick pos=left,
    legend style={
        font=\small,
        inner sep=2pt,
        row sep=1pt,
        legend cell align=left,
    },
    legend image code/.code={
        \draw[draw=none] (0cm,-0.08cm) rectangle (0.3cm,0.18cm);
    },
]

% ---- Grid anchors ----
\addplot[
    ybar,
    fill=blue!100!red!30,
    draw=blue!100!red!30,
    opacity=1.0
] coordinates {
    (0.4297, 2)
    (0.8593, 24)
    (1.2890, 29)
    (1.7186, 5)
    (2.1483, 10)
    (2.5780, 7)
    (3.0076, 1)
    (3.4373, 2)
    (3.8669, 5)
    (4.2966, 7)
    (4.7263, 3)
    (5.1559, 4)
    (5.5856, 3)
    (6.0152, 4)
    (6.4449, 3)
    (6.8746, 1)
    (7.3042, 2)
};

% ---- Random anchors ----
\addplot[
    ybar,
    fill=red!60,
    draw=none,
    opacity=0.6
] coordinates {
    (0.4297, 11)
    (0.8593, 42)
    (1.2890, 24)
    (1.7186, 8)
    (2.1483, 6)
    (2.5780, 5)
    (3.0076, 4)
    (3.4373, 10)
    (3.8669, 2)
};

\legend{Grid Anchors, Random Anchors}
\node[
    anchor=north east,
    align=left,
    font=\small,
    fill=white,
    draw=black!100,
    rounded corners=2pt,
    inner sep=3pt
] at (axis description cs:0.98,0.75) {
    Average Distance (Grid): 2.2801 voxels\\
    Average Distance (Random): 1.2467 voxels
};

\end{axis}
\end{tikzpicture}
}
\caption{Histogram of distances from anchor patch centers to the nearest GM boundary voxel. Our proposed anchor selection method (red bars) produces anchor patches significantly closer to the GM surface than anchors selected on a set grid (blue bars, the baseline). This indicates that randomly selected anchor patches reduce spatial bias and achieve stronger conformity to anatomy compared to the baseline.}
\label{fig:anchorhistogram}
\end{figure}
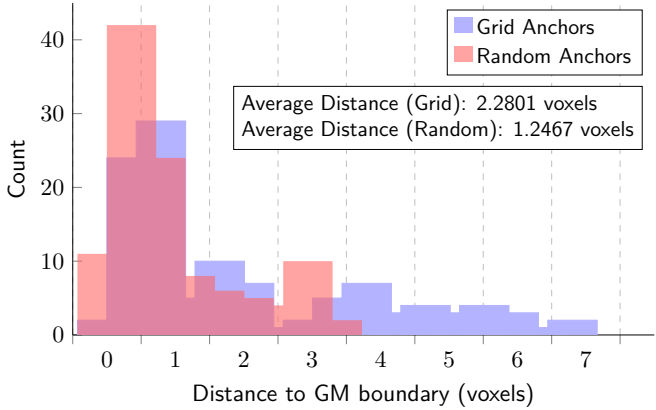

\noindent\textit{Preliminaries:} $\Omega \subset \mathbb{Z}^3$ denotes a discrete 3D brain volume in MNI space with spatial dimensions: 
\begin{equation}
\Omega = \{0, \ldots, X-1\} \times \{0,\ldots,Y-1\} \times \{0,\ldots,Z-1\}.
\end{equation} 

\noindent $M : \Omega \rightarrow \{0,1\}$ is a GM mask, where $M(\mathbf{v}) = 1$ indicates that voxel $\mathbf{v} \in \Omega$ belongs to GM. Cubic patches of side length $s \in \mathbb{N}$ are defined by their top-left coordinate: 
\begin{equation}
\mathbf{t} = (t_x, t_y, t_z) \in \Omega,
\end{equation}

\noindent with the associated voxel set: 
\begin{equation}
\mathcal{P}(\mathbf{t}) = 
\left\{ 
\begin{aligned} 
(x,y,z) \in \Omega \;|\;& 
t_x \le x < t_x + s,\\ 
& t_y \le y < t_y + s,\\ 
& t_z \le z < t_z + s 
\end{aligned} \right\}. 
\end{equation}

We define the GM support of a patch as the functionally relevant tissue a patch actually contains. We quantify this as:
\begin{equation}
G(\mathbf{t}) = \sum_ {v \in \mathcal{P}(\textbf{t})} M(\mathbf{v}).   
\end{equation}

\noindent A patch is considered valid if $G(\mathbf{t}) \geq \tau$, where $\tau > 0$ is a predefined minimum GM overlap threshold. For this research, a $\tau$ of 100 was used, meaning that for a patch to be considered valid, it must contain at least 100 GM voxels. The geometric center of a patch is given as: 
\begin{equation}
\mathbf{c}(\mathbf{t})=\mathbf{t} + (\frac{s}{2}, (\frac{s}{2}, (\frac{s}{2}).
\end{equation}

$\mathcal{A} = \{\mathcal{P}(\mathbf{t}_j)\}_{j=1}^{H}$ denotes a set of anchor patches, which are used as fixed reference regions for defining functional relations like correlations with sampled brain patches. $\mathcal{A}$ is considered valid if each patch is spatially contained within $\Omega$ and satisfies the GM constraint $G(\mathbf{t}_j) \geq \tau$.

\vspace{0.5em}

\noindent\textit{Proposed Method:} In our proposed anchor selection mechanism, anchor patches are generated via random sampling. Specifically, anchor top-left coordinates are sampled from a discrete uniform distribution:
\begin{equation}
    \mathbf{t} \sim \text{Unif}([0, X - s] \times [0, Y -s] \times [0, Z - s]),
\end{equation}
subject to the GM constraint $G(\textbf{t}) \geq \tau$. Sampling continues until a predefined number $H$ of valid anchors ($\mathcal{A_\text{rand}}$) is obtained.
Our proposed method differs from~\cite{liu2024randomizing} in the anchor patch selection process (grid-based vs random). Fig.~\ref{fig:anchorselection} highlights the differences between the two approaches. 
\begin{equation}
\text{ROI} = [(x_\text{min}, x_\text{max}) \times (y_\text{min}, y_\text{max}) \times (z_\text{min}, z_\text{max})].
\end{equation}

\noindent Given a stride vector $\mathbf{\delta} = (\delta_x, \delta_y, \delta_z)$, they construct candidate anchor locations as:
\begin{equation}
\mathbf{t}_{ijk} = (x_\text{min} + i\delta_x + o_x, y_\text{min} + j\delta_y + o_y, z_\text{min} + k\delta_z + o_z),
\end{equation}

\noindent where $i, j, k \in \mathbb{N}$ index the grid and ($o_x, o_y, o_z$) are deterministic offsets chosen to approximately center the grid within the ROI. The resulting anchor set, $\mathcal{A_{\text{grid}}}$, forms an equidistant tiling of the ROI up to boundary effects.

This approach, while reasonably effective, limits models in terms of flexibility and adaptability because the same grid is used for every subject in a dataset, imposing a structural bias and a disregard for individual specificity, as a subject may have functionally distinct regions that do not align well with predefined anchor patches. Additionally, selecting anchor patches in this ROI, grid-based manner poses the risk of many patch centers falling outside of the GM region of the brain (as demonstrated in Fig.~\ref{fig:anchorpatchcenters}(a). For FCA, this poses a problem, as the goal should be to extract the most functionally relevant patches from the GM. 

Our randomized approach to anchor selection fixes this (see Fig.~\ref{fig:anchorpatchcenters})(b), offering anchor patches that are more diverse across patients, as well as more conform to brain anatomy. This ensures that more functionally relevant patches are extracted from the anchor regions. The better anatomical conformity is validated by Fig.~\ref{fig:anchorhistogram}. As seen, our proposed anchor selection approach produces anchor patches whose centers are 45.32\% closer to the GM boundary compared to those produced by the baseline approach.

\subsection{Multi-Scale Patch Sampling}
\label{subsec:iterative}

\noindent\textit{Preliminaries:} rs-fMRI data is represented as a 4D array $\mathbf{X \in \mathbb{R}^{T \times X \times Y \times Z}}$, where $\mathbf{X}_t(\mathbf{v})$ denotes the BOLD signal at time $t \in \{1, \dots, T\}$ and voxel $\mathbf{v} \in \Omega$. All volumes are assumed to be in the same MNI coordinate system as the previously selected anchor patches. $L : \Omega \rightarrow \{0, 1, \dots, H\}$ denotes the anchor label image, where $L(\mathbf{v}) = j$ indicates that voxel $\mathbf{v}$ belongs to anchor patch $j$, and $L(\mathbf{v}) = 0$ denotes background. For anchor $j$, the GM-restricted support is defined as:
\begin{equation}
\mathcal{A}_j = {\mathbf{v} \in \Omega | L(\mathbf{v}) = j}.
\end{equation}

For a patch side length $s$, a sampled patch is indexed by its center coordinate:
\begin{equation}
    \mathbf{p} = (p_x, p_y, p_z) \in \Omega,
\end{equation}

\noindent which induces a cubic voxel set:
\begin{equation}
\mathcal{P}(\mathbf{p}; s) =
\left\{ 
\begin{aligned} 
\mathbf{v} \in \Omega | |v_x - p_x| < \frac{s}{2},\\
|v_y - p_y| < \frac{s}{2},\\
|v_z - p_z| < \frac{s}{2}
\end{aligned} \right\}. 
\end{equation}

\noindent with boundary clipping implicitly enforced by the condition $\mathbf{v} \in \Omega$. A sampled patch is considered valid if its GM support satisfies $G(\mathbf{p}; s) \geq \theta$. In this research, the threshold is set to $\theta = 1$.

For any voxel set $\mathcal{S} \subseteq \Omega$ with nonzero cardinality, the mean BOLD time series is defined as:
\begin{equation}
\mu(\mathcal{S})_t = \frac{1}{|\mathcal{S}|} \sum_{\mathbf{v \in \mathcal{S}}} \mathbf{X}_t(\mathbf{v}).
\end{equation}

%%%%%%%%%%%
\begin{figure}[t]
    \centering
    \resizebox{\linewidth}{!}
    {
    \begin{tabular}{m{0.5\linewidth} m{0.5\linewidth}}
    Random Patch Sampling & Multi-Scale Patch Sampling \\[0.75em]
    \includegraphics[width=\linewidth]{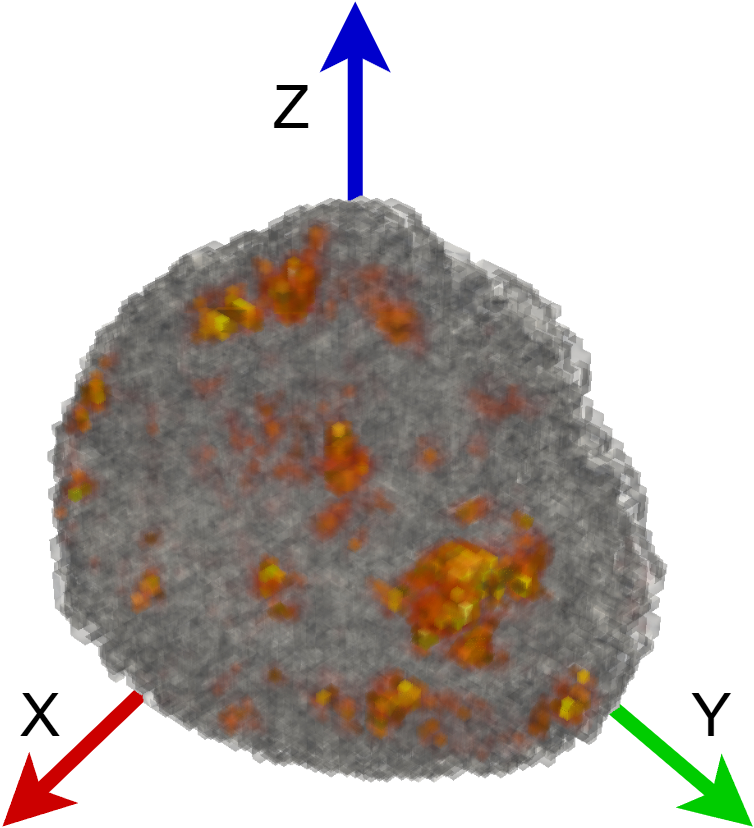} &
    \includegraphics[width=\linewidth]{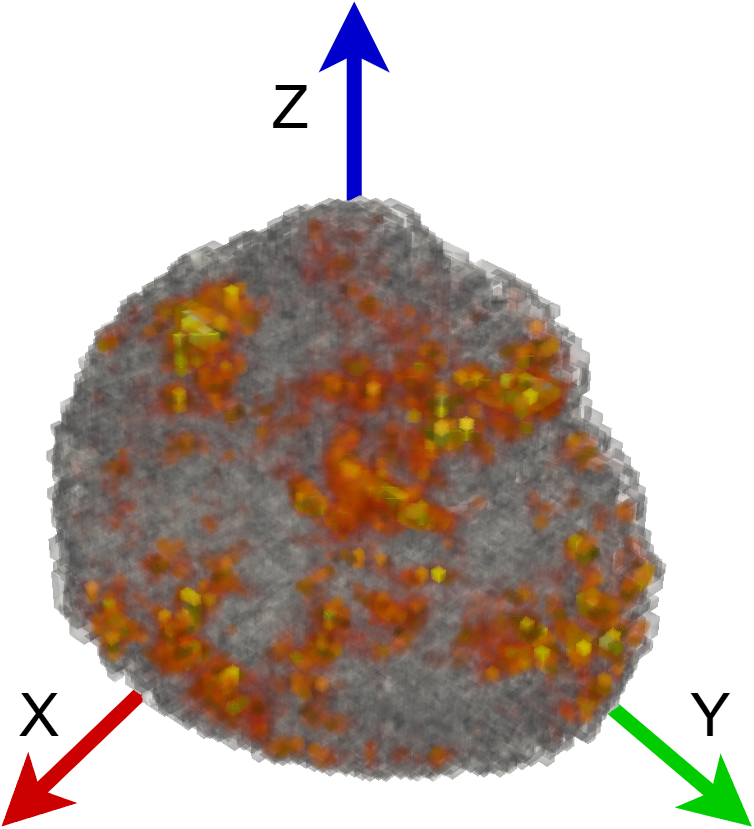}
    \\[7.0em]
    GM Coverage (\%): 66.05 & GM Coverage (\%): 91.73
    \end{tabular}
    }
    \caption{GM coverage of the baseline random patch sampling strategy and the proposed multi-scale patch sampling. Visualized are the centers of the extracted patches, with hotter (indicated by the color yellow) regions in the heatmap indicating areas in the GM captured in multiple patches. Multi-scale patch sampling yields a substantially denser and more uniform coverage ($\uparrow$ 25.68\%) of cortical GM compared to single-pass random sampling.}
    \label{fig:gmcoverage}
\end{figure}
%%%%%%%%%%%%%%

\noindent The anchor time series for anchor $j$ is given by $\mathbf{a}_j = \mu(\mathcal{A_j})$. Similarly, the time series for sampled patch $i$, centered at $\mathbf{p}_i$, is computed over GM only as:
\begin{equation}
\mathbf{b}_i = \mu(\mathcal{P}(\mathbf{p}_i; s) \cap \{\mathbf{v} | M(\mathbf{v}) = 1\}).
\end{equation}

To retain spatial information, the center coordinate of each sampled patch is normalized by the image dimensions, yielding:
\begin{equation}
\tilde{\mathbf{p}}_i =(\frac{p_{ix}}{X},\frac{p_{iy}}{Y},\frac{p_{iz}}{Z})\in [0,1]^3.
\end{equation}
\noindent The position-aware patch representation, $\mathbf{u}$, is formed by concatenating the normalized position with the patch time series:
\begin{equation}
\mathbf{u}_i =([\,\tilde{\mathbf{p}}_i \;\; \mathbf{v}_i\,)]\in \mathbb{R}^{3+T}.
\end{equation}

\begin{figure*}[t]
    \centering
    \includegraphics[width=0.8\linewidth]{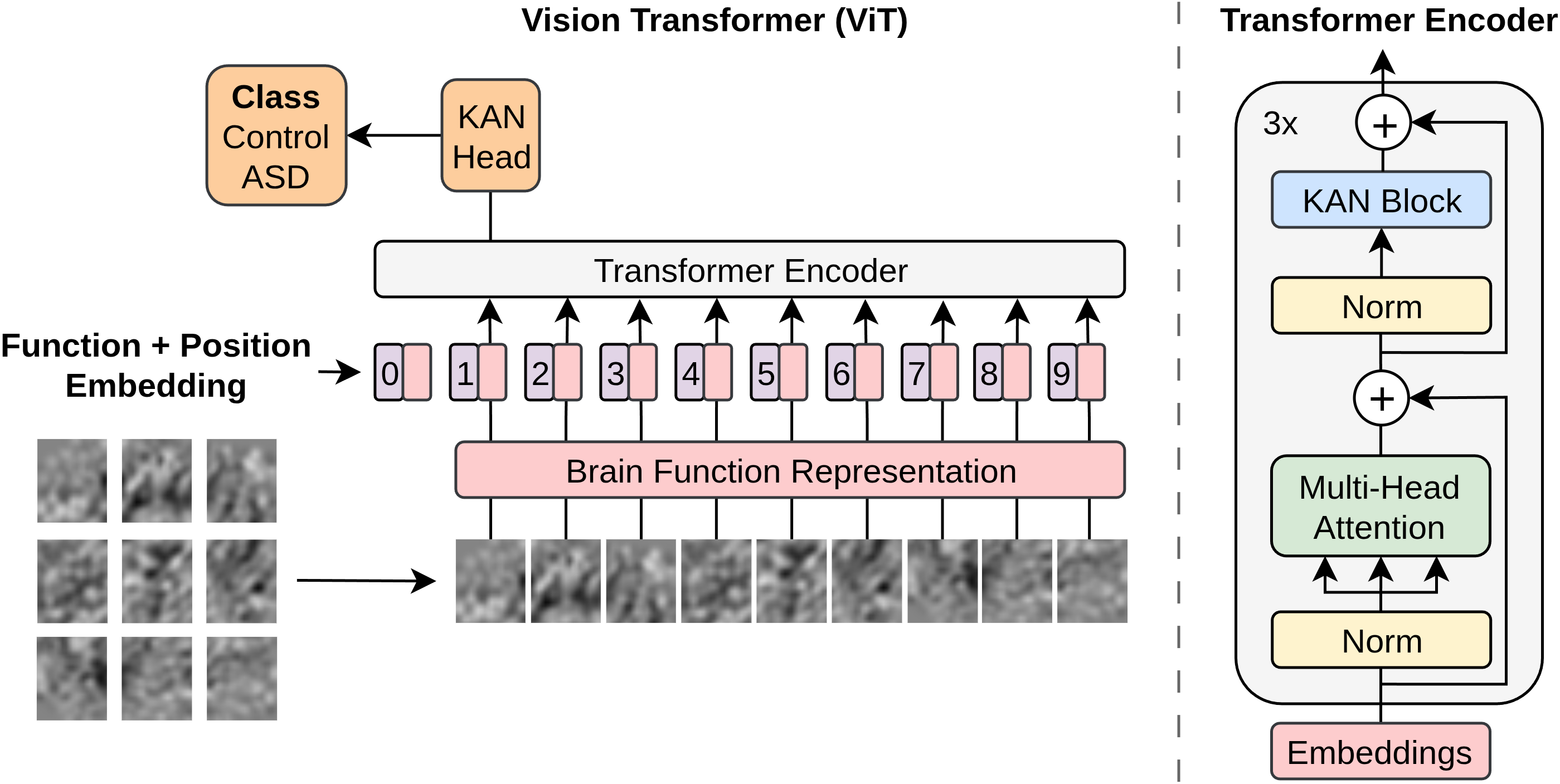}
    \caption{The classification network used in our proposed ABFR-KAN framework is a ViT-style encoder-only network. The transformer is fed patches derived from rs-fMRIs that are embedded with spatial position information and passed through the encoder. The binary classification prediction (control or ASD) is produced by the KAN head. In the encoder, we replace the MLP block with a KAN block.}
    \label{fig:transformer}
\end{figure*}

Given $N$ sampled patches, the patch-to-anchor FC matrix $\mathbf{F} \in \mathbb{R}^{N \times H}$ is defined by:
\begin{equation}
F_{ij} = \mathrm{Corr}(\mathbf{v}_i, \mathbf{a}_j),
\end{equation}
where $\mathrm{Corr}(\cdot,\cdot)$ denotes Pearson correlation. Each row of $\mathbf{F}$ represents the functional relationship between one sampled patch and all anchor patches.

\vspace{0.5em}

\noindent\textit{Proposed Method:} Our proposed patch sampling approach performs multiple independent sampling iterations with different patch sizes. Fig.~\ref{fig:patchsampling} shows a comparison with the baseline method. Specifically, we use $R = 3$ iterations with patch sizes $s_1 = 8$, $s_2 = 12$, and $s_3 = 16$. An ablation study demonstrates the impact of the number of iterations (Fig.~\ref{fig:iterationablation}). In each iteration, 256 valid patches are sampled using the same acceptance rule, producing iteration-specific FC matrices, $\mathbf{F}^{(r)}$, and position-aware representations, $\{\mathbf{u}_i^{(r)}\}$. We obtain the final FC representation by averaging the FC matrices across iterations, while aggregating the position-aware representations via concatenation.

The baseline method \cite{liu2024randomizing} that our proposed approach aims to improve upon simply samples patches once per subject using a fixed patch size of $s = 8$. Candidate patch centers are drawn uniformly from the brain volume and accepted if they satisfy the GM support constraint. Sampling terminates when $N = 256$ valid patches have been obtained. Anchor time series are computed first, followed by the extraction of sampled patch time series and the construction of the FC matrix, $\mathbf{F}$.

Our proposed multi-scale sampling approach enhances the robustness of extracted rs-fMRI features by reducing the impact of single-scale patch selection biases and improving anatomical coverage (as demonstrated in Fig.~\ref{fig:gmcoverage}). Additionally, our proposed approach effectively reduces the sampling variance in FC estimation (Fig.~\ref{fig:iterationablation}). The reduction of this variance should improve feature stability, reduce the risk of overfitting to sampling artifacts, and improve reproducibility between sites.

\subsection{Classification Network}

For our proposed ABFR-KAN approach, we utilized a vision transformer (ViT)-style architecture similar to the one introduced by \cite{dosovitskiy2020image}. The specific ViT architecture is visualized in Fig.~\ref{fig:transformer} and is discussed in detail in this section.

\vspace{0.5em}

\noindent\textit{Inputs:} For each subject, the model receives a set of sampled brain patches represented by their functional relationships to anchor patches and their spatial locations. Specifically, after patch sampling and anchor-based FC computation, each subject is represented by a patch-anchor FC matrix, $\mathbf{F} \in \mathbb{R}^{N \times H}$, where $N$ is the number of sampled patches and $H$ is the number of anchor patches. Each row corresponds to the FC profile of one sampled patch with respect to all anchors. A corresponding set of normalized spatial coordinates, $\{\tilde{\mathbf{p}}_i\}_{i=1}^{N} \subset [0,1]^3$ encodes the spatial location of each sampled patch center. These components jointly characterize both the functional and spatial properties of the sampled patches.

\vspace{0.5em}

\noindent\textit{Patch Selection:} Before entering the transformer encoder, a learnable Top-K pooling operation is applied to reduce the number of patch tokens and emphasize the most informative ones. Each patch feature vector, which is a row of $\mathbf{F}$, $\mathbf{f}_i \in \mathbb{R}^H$, is assigned an importance score through a learnable projection followed by a nonlinearity. Patches are then ranked by these scores, and only the top 80\% of patches are retained. The same selection is applied consistently to both the functional features and their corresponding spatial coordinates. This operation encourages the model to focus on functionally salient patches that are most discriminative for the classification task.

\vspace{0.5em}

\noindent\textit{Positional Encoding:} To retain spatial information after Top-K pooling, each retained patch’s normalized spatial coordinate is projected into the same embedding space as the functional features using a learnable linear transformation. The resulting positional embedding is added to the patch’s functional feature vector. This additive fusion allows the transformer to jointly reason about FC patterns and spatial context, without imposing any predefined spatial ordering or grid structure \cite{dosovitskiy2020image}.

\vspace{0.5em}

\noindent\textit{Transformer Encoder:} The resulting set of patch embeddings is processed by a multi-layer transformer encoder composed of stacked self-attention and feedforward blocks. Within each self-attention layer, patches are treated as tokens and allowed to attend to one another. In contrast to pairwise FC, self-attention captures global dependencies across all sampled patches simultaneously, allowing the network to model distributed functional patterns associated with ASD. Each transformer layer follows a standard pre-normalization design in that layer normalization is applied before both the attention and feedforward blocks, and residual connections are used to ensure stable gradient flow and preserve information across layers. Through successive layers, patch representations are progressively refined, integrating information from across the brain.

\vspace{0.5em}

\noindent\textit{Global Aggregation and Classification:} After being passed through the encoder, the patch-level representations are aggregated into a single subject-level embedding by mean pooling across the patch dimension. This pooling operation produces a fixed-length representation regardless of the number of retained patches. The aggregated embedding is then passed through a lightweight classification head consisting of layer normalization followed by a linear classifier. The output is a two-dimensional logit vector corresponding to the ASD and control classes.

%%%%%%%%%%%
\begin{figure}[t]
    \centering
    \resizebox{\linewidth}{!}{
    \begin{tabular}{m{0.37\linewidth} m{0.37\linewidth} m{0.25\linewidth}}
    \includegraphics[width=\linewidth]{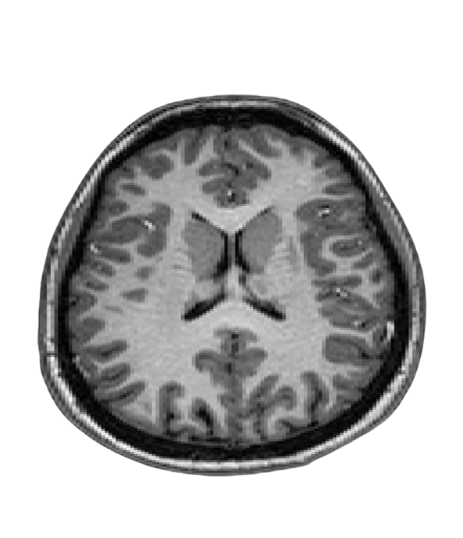} &
    \includegraphics[width=\linewidth]{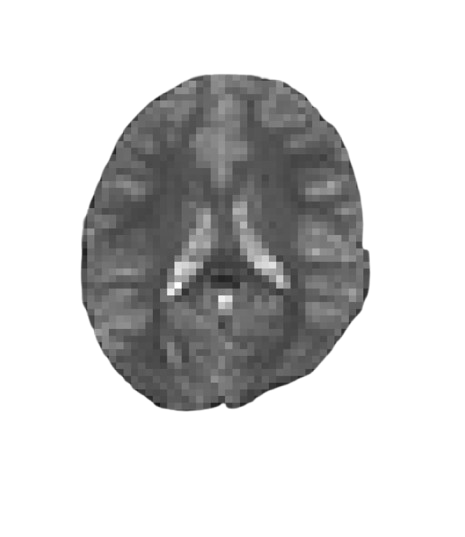} &
    \includegraphics[width=\linewidth]{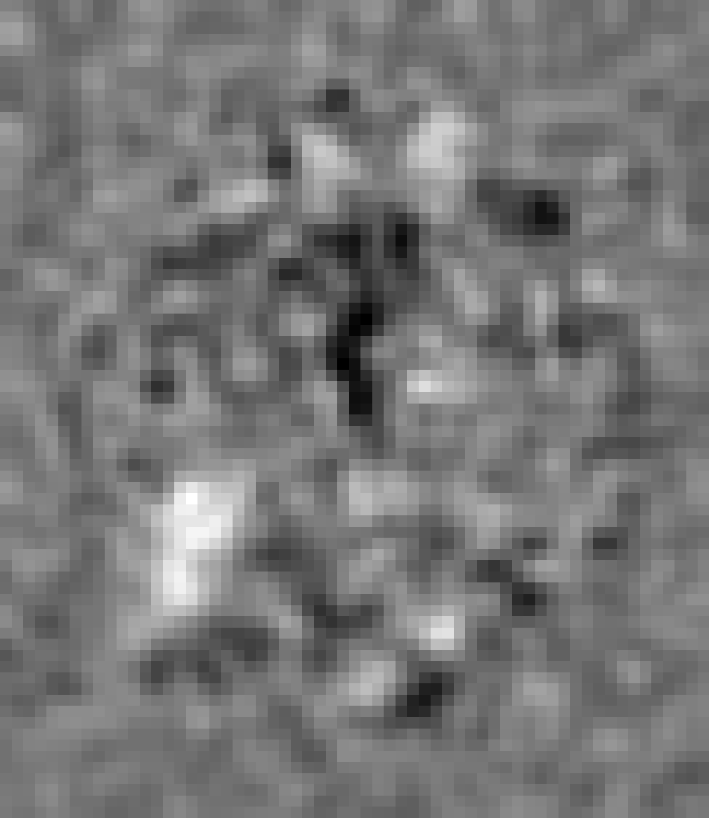} \\
    \parbox[t]{\linewidth}{\centering\large
        Magnetization-\\
        Prepared Rapid\\
        Gradient Echo\\
        (MP-RAGE)
    } &
    \parbox[t]{\linewidth}{\centering
        rs-fMRI
    } &
    \parbox[t]{\linewidth}{\centering
        DPARSF pre-\\
        processed rs-\\
        fMRI patch
    }
    \end{tabular}
    }
    \caption{Raw and preprocessed data for a single patient in the ABIDE I dataset.}
    \label{fig:data}
\end{figure}
%%%%%%%%%%%%%%
 
\section{Experimental Evaluation}

\begin{table*}[t]
    \centering
    \caption{Performance comparison of ABFR-KAN against state-of-the-art methods on the NYU and UM sites of the ABIDE I dataset and the KKI site of the ABIDE II dataset. Data samples were obtained via our proposed random anchor selection and multi-scale patch sampling strategy. Mean±stdev scores are reported by averaging the results of each fold. Best and second-best results are \textbf{bolded} and \underline{underlined}, respectively. ($\dagger$) indicates baselines that ABFR-KAN improves upon ACC with statistical significance.}
    \vspace{0.5em}
    \resizebox{\linewidth}{!}{
    \begin{tabular}{lcccccc}
         \toprule
         \rowcolor{gray!25}
         \multicolumn{7}{c}{\textbf{\large NYU Site Classification Performance}} \\
         \midrule
         Method & ACC & AUC & F1 & PRE & SEN & SPE \\
         \midrule
         SVM ($\dagger$) & 0.5615 ± 0.0301 & 0.4740 ± 0.1130 & 0.6750 ± 0.0237 & 0.5888 ± 0.0317 & 0.7974 ± 0.0681 & 0.2495 ± 0.1005 \\
         BrainNetCNN ($\dagger$) & 0.5438 ± 0.0390 & 0.5151 ± 0.0464 & 0.6527 ± 0.0305 & 0.5790 ± 0.0211 & 0.7495 ± 0.0550 & 0.2689 ± 0.0508 \\
         GAT ($\dagger$) & 0.5545 ± 0.0107 & 0.5500 ± 0.0115 & 0.6556 ± 0.0118 & 0.5894 ± 0.0181 & 0.7414 ± 0.0426 & 0.3061 ± 0.0490 \\
         GCN ($\dagger$) & 0.5758 ± 0.0336 & 0.5731 ± 0.0364 & 0.6806 ± 0.0290 & 0.5984 ± 0.0301 & 0.7896 ± 0.0300 & 0.2888 ± 0.0405 \\
         BrainGNN ($\dagger$) & 0.5731 ± 0.0132 & 0.5571 ± 0.0238 & 0.7285 ± 0.0106 & 0.5731 ± 0.0132 & \textbf{1.0000 ± 0.0000} & 0.0000 ± 0.0000 \\
         MVS-GCN ($\dagger$) & 0.6376 ± 0.0811 & 0.6163 ± 0.0746 & 0.7563 ± 0.0403 & 0.6269 ± 0.0739 & \underline{0.9700 ± 0.0600} & 0.1990 ± 0.2282 \\
         KD-Transformer ($\dagger$) & 0.6610 ± 0.0272 & 0.6137 ± 0.0692 & 0.7225 ± 0.0216 & 0.6942 ± 0.0830 & 0.7768 ± 0.1014 & 0.6437 ± 0.0438 \\
         RandomFR & \underline{0.7079 ± 0.0388} & \underline{0.6544 ± 0.0583} & \underline{0.7734 ± 0.0379} & \underline{0.6986 ± 0.0435} & 0.8784 ± 0.1024 & \underline{0.6783 ± 0.0486} \\
         \rowcolor{green!17.5}
         \midrule
         ABFR-KAN ($\dagger$) & \textbf{0.7427 ± 0.0342} & \textbf{0.7041 ± 0.0708} & \textbf{0.8003 ± 0.0200} & \textbf{0.7267 ± 0.0469} & 0.8984 ± 0.0646 & \textbf{0.7159 ± 0.0434} \\
         \midrule
         \rowcolor{gray!25}
         \multicolumn{7}{c}{\textbf{\large UM Site Classification Performance}} \\
         \midrule
         Method & ACC & AUC & F1 & PRE & SEN & SPE \\
         \midrule
         SVM ($\dagger$) & 0.6000 ± 0.0603 & 0.4866 ± 0.1278 & 0.6978 ± 0.0490 & 0.5655 ± 0.0834 & 0.9385 ± 0.0754 & 0.2802 ± 0.0971 \\
         BrainNetCNN ($\dagger$) & 0.5362 ± 0.0379 & 0.5563 ± 0.0526 & 0.5378 ± 0.0549 & 0.5460 ± 0.1043 & 0.5552 ± 0.0996 & 0.5404 ± 0.0998 \\
         GAT ($\dagger$) & 0.5435 ± 0.0422 & 0.5741 ± 0.0556 & 0.6042 ± 0.0481 & 0.5405 ± 0.0927 & 0.7108 ± 0.0846 & 0.3981 ± 0.0758 \\
         GCN ($\dagger$) & 0.5601 ± 0.0462 & 0.5898 ± 0.0579 & 0.5714 ± 0.0657 & 0.5609 ± 0.0792 & 0.6085 ± 0.1320 & 0.5337 ± 0.0557 \\
         BrainGNN ($\dagger$) & 0.4392 ± 0.0563 & 0.5521 ± 0.0365 & 0.3502 ± 0.2860 & 0.2722 ± 0.2275 & 0.5246 ± 0.4440 & 0.4964 ± 0.4393 \\
         MVS-GCN ($\dagger$) & 0.6727 ± 0.0445 & 0.6357 ± 0.0635 & 0.6953 ± 0.0571 & 0.6498 ± 0.0667 & 0.7744 ± 0.1488 & 0.5509 ± 0.2138 \\
         KD-Transformer ($\dagger$) & 0.6727 ± 0.0182 & \underline{0.6486 ± 0.0464} & \underline{0.7233 ± 0.0356} & 0.6194 ± 0.0487 & \textbf{0.8730 ± 0.0367} & 0.6674 ± 0.0381 \\
         RandomFR ($\dagger$) & \underline{0.7182 ± 0.0530} & 0.6426 ± 0.1395 & 0.6999 ± 0.1311 & \underline{0.7169 ± 0.1104} & 0.7397 ± 0.2197 & \underline{0.6984 ± 0.0780} \\
         \rowcolor{green!17.5}
         \midrule
         ABFR-KAN & \textbf{0.7727 ± 0.0287} & \textbf{0.7184 ± 0.0620} & \textbf{0.7682 ± 0.0482} & \textbf{0.7840 ± 0.1086} & \underline{0.7837 ± 0.1314} & \textbf{0.7768 ± 0.0270} \\
         \midrule
         \rowcolor{gray!25}
         \multicolumn{7}{c}{\textbf{\large KKI Site Classification Performance}} \\
         \midrule
         Method & ACC & AUC & F1 & PRE & SEN & SPE \\
         \midrule
         SVM ($\dagger$) & 0.6182 ± 0.0680 & 0.4538 ± 0.1260 & 0.0800 ± 0.1600 & 0.2000 ± 0.4000 & 0.0500 ± 0.1000 & \textbf{1.0000 ± 0.0000} \\
         BrainNetCNN ($\dagger$) & 0.4973 ± 0.0884 & 0.4358 ± 0.1379 & 0.2801 ± 0.1082 & 0.3455 ± 0.1553 & 0.2429 ± 0.0943 & 0.6714 ± 0.1057 \\
         GAT ($\dagger$) & 0.5756 ± 0.0746 & 0.5593 ± 0.0906 & 0.3496 ± 0.0883 & 0.4795 ± 0.1163 & 0.2972 ± 0.1098 & 0.7673 ± 0.1339 \\
         GCN ($\dagger$) & 0.5818 ± 0.0533 & 0.4717 ± 0.1318 & 0.1344 ± 0.1231 & 0.3609 ± 0.2960 & 0.0835 ± 0.0789 & \underline{0.9140 ± 0.0381} \\
         BrainGNN ($\dagger$) & 0.6000 ± 0.0445 & 0.4974 ± 0.0357 & 0.0000 ± 0.0000 & 0.0000 ± 0.0000 & 0.0000 ± 0.0000 & \textbf{1.0000 ± 0.0000} \\
         MVS-GCN ($\dagger$) & 0.6000 ± 0.0445 & 0.4914 ± 0.1078 & 0.0000 ± 0.0000 & 0.0000 ± 0.0000 & 0.0000 ± 0.0000 & \textbf{1.0000 ± 0.0000} \\
         KD-Transformer ($\dagger$) & 0.6364 ± 0.0575 & 0.3757 ± 0.0642 & 0.2133 ± 0.1759 & 0.5000 ± 0.4472 & 0.1400 ± 0.1158 & 0.5557 ± 0.0509 \\
         RandomFR & \underline{0.7455 ± 0.0600} & \underline{0.6248 ± 0.1552} & \underline{0.6457 ± 0.1855} & \underline{0.7076 ± 0.1774} & \underline{0.6200 ± 0.2159} & 0.7148 ± 0.1162 \\
         \rowcolor{green!17.5}
         \midrule
         ABFR-KAN & \textbf{0.7636 ± 0.0445} & \textbf{0.6505 ± 0.0869} & \textbf{0.6588 ± 0.1329} & \textbf{0.8033 ± 0.1675} & \textbf{0.6300 ± 0.2015} & 0.7388 ± 0.0648 \\
         \bottomrule
    \end{tabular}
    }
    \label{tab:singlesiteresults}
\end{table*}

\begin{table*}[t]
    \centering
    \caption{Cross-site classification performance of ABFR-KAN compared to SOTA baselines. The data were extracted via our proposed random anchor selection and multi-scale patch sampling strategy. Mean±stdev scores are reported by averaging the results of five runs. Best and second-best results are \textbf{bolded} and \underline{underlined}, respectively. ($\dagger$) indicates baselines that ABFR-KAN improves upon the ACC in a statistically significant manner.}
    \vspace{0.5em}
    \resizebox{\linewidth}{!}{
    \begin{tabular}{lcccccc}
         \toprule
         \rowcolor{gray!25}
         \multicolumn{7}{c}{\textbf{\large Classification Performance (Train: NYU, Test: UM)}} \\
         \midrule
         Method & ACC & AUC & F1 & PRE & SEN & SPE \\
         \midrule
         SVM & \underline{0.5527 ± 0.0260} & 0.5170 ± 0.0706 & 0.5623 ± 0.0582 & 0.5479 ± 0.0204 & 0.5855 ± 0.1050 & 0.5200 ± 0.0676 \\
         BrainNetCNN ($\dagger$) & 0.5055 ± 0.0473 & 0.5067 ± 0.0452 & 0.5413 ± 0.1714 & 0.4850 ± 0.0588 & 0.6764 ± 0.2857 & 0.3345 ± 0.2336 \\
         GAT ($\dagger$) & 0.4909 ± 0.0359 & 0.5070 ± 0.0468 & 0.3322 ± 0.2777 & 0.3817 ± 0.1562 & 0.4145 ± 0.4223 & 0.5673 ± 0.3718 \\
         GCN ($\dagger$) & 0.5073 ± 0.0253 & 0.4976 ± 0.0397 & 0.2652 ± 0.2633 & 0.5101 ± 0.3165 & 0.2836 ± 0.3200 & \textbf{0.7309 ± 0.2801} \\
         BrainGNN ($\dagger$) & 0.5000 ± 0.0000 & 0.5869 ± 0.0093 & 0.6667 ± 0.0000 & 0.5000 ± 0.0000 & \textbf{1.0000 ± 0.0000} & 0.0000 ± 0.0000 \\
         MVS-GCN ($\dagger$) & 0.4945 ± 0.0159 & 0.5498 ± 0.0668 & 0.5682 ± 0.1883 & 0.4788 ± 0.0450 & \underline{0.8073 ± 0.3418} & 0.1818 ± 0.3120 \\
         KD-Transformer ($\dagger$) & 0.5164 ± 0.0278 & 0.5035 ± 0.0619 & 0.3717 ± 0.2104 & 0.5029 ± 0.1373 & 0.3673 ± 0.2705 & \underline{0.6655 ± 0.2463} \\
         RandomFR & \textbf{0.6109 ± 0.0668} & \underline{0.6364 ± 0.0673} & \underline{0.7224 ± 0.0379} & \underline{0.6840 ± 0.0521} & 0.7689 ± 0.0647 & 0.5662 ± 0.0714\\
         \rowcolor{green!17.5}
         \midrule
         ABFR-KAN & 0.5517 ± 0.0462 & \textbf{0.7181 ± 0.0619} & \textbf{0.7473 ± 0.0320} & \textbf{0.7036 ± 0.0489} & 0.8021 ± 0.0598 & 0.6147 ± 0.0632\\
         \midrule
         \rowcolor{gray!25}
         \multicolumn{7}{c}{\textbf{\large Classification Performance (Train: UM, Test: NYU)}} \\
         \midrule
         Method & ACC & AUC & F1 & PRE & SEN & SPE \\
         \midrule
         SVM ($\dagger$) & 0.5427 ± 0.0563 & 0.4909 ± 0.0618 & 0.5827 ± 0.1626 & 0.5846 ± 0.0365 & 0.6286 ± 0.2331 & 0.4274 ± 0.1875 \\
         BrainNetCNN ($\dagger$) & 0.4585 ± 0.0413 & 0.4805 ± 0.0402 & 0.2753 ± 0.1844 & 0.5343 ± 0.1367 & 0.2204 ± 0.1977 & \textbf{0.7781 ± 0.1785} \\
         GAT ($\dagger$) & 0.5392 ± 0.0183 & 0.4872 ± 0.0210 & \underline{0.6758 ± 0.0298} & 0.5654 ± 0.0061 & \underline{0.8449 ± 0.0883} & 0.1288 ± 0.0834 \\
         GCN ($\dagger$) & \underline{0.5556 ± 0.0414} & 0.4979 ± 0.0243 & \textbf{0.6931 ± 0.0574} & 0.5696 ± 0.0176 & \textbf{0.8939 ± 0.1327} & 0.1014 ± 0.0816 \\
         BrainGNN ($\dagger$) & 0.4643 ± 0.0530 & \underline{0.5703 ± 0.0128} & 0.2212 ± 0.2616 & 0.3561 ± 0.2925 & 0.2347 ± 0.3561 & \underline{0.7726 ± 0.3551} \\
         MVS-GCN ($\dagger$) & 0.4795 ± 0.0409 & 0.5420 ± 0.0252 & 0.3473 ± 0.1927 & \underline{0.5887 ± 0.0392} & 0.3122 ± 0.2825 & 0.7041 ± 0.2939 \\
         KD-Transformer ($\dagger$) & 0.5181 ± 0.0530 & 0.5001 ± 0.0531 & 0.5617 ± 0.1255 & 0.5747 ± 0.0464 & 0.5939 ± 0.2410 & 0.4164 ± 0.2306 \\
         RandomFR ($\dagger$) & 0.5252 ± 0.0330 & 0.4866 ± 0.1155 & 0.6064 ± 0.0159 & 0.5873 ± 0.0445 & 0.6279 ± 0.0512 & 0.4628 ± 0.0603\\
         \rowcolor{green!17.5}
         \midrule
         ABFR-KAN & \textbf{0.5707 ± 0.0147} & \textbf{0.5714 ± 0.0500} & 0.6752 ± 0.0322 & \textbf{0.6128 ± 0.0397} & 0.7536 ± 0.0551 & 0.5019 ± 0.0584\\
         \bottomrule
    \end{tabular}
    }
    \label{tab:crosssiteresults}
\end{table*}

\subsection{Data}
\label{subsec:data}
We evaluated our proposed ABFR-KAN using pre-processed neuroimaging data from the Autism Brain Imaging Data Exchange (ABIDE) \cite{craddock2013neuro, di2014autism} I and II. \\
\textbf{ABIDE I:} The preprocessed ABIDE I repository contains data collected from a total of 1,112 patients at various sites. In this work, we trained and evaluated our ABFR-KAN approach on data from the New York University (NYU) Langone Medical Center and the University of Michigan (UM) Functional MRI Center. Both sites were processed using the Data Processing Assistant for Resting-State fMRI (DPARSF) \cite{yan2010dparsf} method. In total, 64 male (age range: 7-39) and 9 female (age range: 10-38) patients with ASD diagnoses were selected from the NYU site, along with 72 male (age range: 6-31) and 26 female (age range: 8-29) patients from the control group. From the UM site, we selected a balanced dataset consisting of information from 55 patients in the control group (46 male, aged 8 to 18; 9 female, aged 9 to 18) and 55 in the ASD group (38 male, aged 8 to 18; 17 female, aged 9 to 19). 

\noindent \textbf{ABIDE II:} The ABIDE II repository contains data from a total of 1,114 patients, collected from 19 sites. In this work, we select data from the largest ABIDE II site, the Kennedy Krieger Institute (KKI). Preprocessing on this site is consistent with that of the NYU and UM sites, which has already been discussed. In this data subset, there were a total of 211 patients, with an age range of 8-13 years old. This subset represents the most imbalanced of the data subsets we explore in this work, with 56 patients having an ASD diagnosis, and 155 belonging to the control group.

Within each data subset, ASD diagnoses were made using a combination of clinical judgment and standardized ``gold standard'' instruments, specifically the Autism Diagnostic Observation Schedule (ADOS) \cite{lord2000autism} and the Autism Diagnostic Interview--Revised (ADI-R) \cite{rutter2003autism}, with the diagnostic categories based on the Diagnostic and Statistical Manual of Mental Disorders, Fourth Edition, Text Revision (DSM-IV-TR) \cite{APA2000DSMIVTR} criteria. The bias towards the male sex that is shown in both data sites is consistent with ASD literature, where it has been shown that males are about four times as likely to be diagnosed with ASD compared to females \cite{leow2024towards}. A visualization of the data types extracted from the ABIDE I dataset is shown in Fig.~\ref{fig:data}.

\subsection{Implementation Details}
\textit{Machine Configuration:} The ABFR-KAN model was implemented in PyTorch and trained on an Intel (R) Xeon (R) w7-2475X, 2600MHz machine with dual NVIDIA A4000X2 GPUs (32GB). 

\vspace{0.5em}

\noindent\textit{Baselines:} We compare our proposed ABFR-KAN against state-of-the-art (SOTA) models for brain disorder diagnosis. These include a support vector machine (SVM) \cite{cortes1995support}; a CNN-based method in BrainNetCNN \cite{kawahara2017brainnetcnn}; a graph attention network (GAT) \cite{velivckovic2018graph}; GCN-based methods in the original GCN \cite{qin2022using} and MVS-GCN \cite{wen2022mvs}; a GNN-based approach in BrainGNN \cite{li2021braingnn}, and transformer-based approaches in KD-Transformer \cite{zhang2022diffusion} and RandomFR \cite{liu2024randomizing}.

\vspace{0.5em}

\noindent\textit{Model Backbones:} We investigate the performance of ABFR-KAN under multiple backbones for the classification network. Specifically, we experiment using a CNN-based backbone in EfficientNetV2 \cite{tan2021efficientnetv2}, transformer-based backbones using a ViT~\cite{dosovitskiy2020image} and Data-efficient image Transformer (DeiT)~\cite{touvron2021training}, and a Mamba-based backbone in Vision Mamba (Vim)~\cite{zhu2024vision}.

\vspace{0.5em}

\noindent\textit{KAN Variants:} We explored five different KAN variants in our ABFR-KAN framework: Efficient-KAN \cite{blealtan2024efficient}, FastKAN \cite{li2024kolmogorov}, FasterKAN \cite{delis2024fasterkan}, Wav-KAN \cite{bozorgasl2405wav}, and ChebyKAN \cite{guo2024chebykan}. Efficient-KAN, FastKAN, and FasterKAN all utilize spline-based basis functions, Wav-KAN utilizes fixed wavelet transforms, and ChebyKAN uses truncated Chebyshev polynomial expansions on normalized inputs.

\noindent\textit{Training:} A 5-fold cross-validation strategy was used to assess the model's performance. On the NYU site, the four folds used for training contained 34 samples each, while the fold kept out of the training process contained 35 samples. For the UM site, the split was 22 samples in each fold, and on KKI, there were 42 samples in the folds used for training and 43 samples in the one kept out of training. For classification, we minimize the cross-entropy loss. The model was trained for 100 epochs, using the AdamW optimizer with a learning rate of 0.001. We used cosine annealing with warm restarts as the learning rate scheduler, along with a weight decay of 0.0001. 

\vspace{0.5em}

\noindent\textit{Evaluation:} Model performance is gauged using traditional metrics for classification tasks, namely accuracy (ACC), area under the curve (AUC), F1 score (F1), precision (PRE), sensitivity (SEN), and specificity (SPE). For the non-AUC metrics, class labels are obtained by applying an argmax over the softmax output, corresponding to a decision threshold of 0.5 in the binary setting. For AUC computation, we use the continuous softmax probability of the positive class rather than binarized predictions, allowing evaluation across all possible decision thresholds.

\subsection{Results}

\subsubsection{Single-Site Performance}
\label{subsubsec:singlesiteresults}

Table~\ref{tab:singlesiteresults} shows the performance of our proposed ABFR-KAN compared to SOTA baselines on the NYU and UM sites of the ABIDE I dataset. In the table, ABFR-KAN is represented by the best-performing KAN variation and configuration as determined by ablation experiments we performed (See Section~\ref{subsubsec:ablation}). For the NYU site, this means that the FastKAN variant was used in a KAN-KAN configuration, where FastKAN replaced the MLP components in both the ViT encoder and the classification head. For the UM site, ABFR-KAN is customized as the WavKAN variant, using an MLP-KAN configuration, where WavKAN replaces only the MLP classification head. Each of the compared-against baseline methods was trained and evaluated on the same data (that is, data that had been processed using our proposed random anchor selection and multi-scale patch sampling approach) as ABFR-KAN, to ensure fair comparison.

From the results, it is clear that ABFR-KAN consistently outperforms both the direct strong baseline in RandomFR, as well as all other compared baseline methods. When trained on the NYU site, ABFR-KAN outperforms RandomFR in terms of ACC, AUC, F1, PRE, SEN, and SPE by 3.48\%, 4.97\%, 2.69\%, 2.81\%, 2.00\%, and 3.76\%, respectively. When trained on the UM site, these improvements become 5.45\%, 7.58\%, 6.83\%, 6.71\%, 4.40\%, and 7.84\%, respectively. On the KKI site, the improvements are by 1.81\%, 2.57\%, 1.31\%, 9.57\%, 1.00\%, and 2.40\%, respectively. On the third and fourth best-performing baselines (KD-Transformer and MVS-GCN, respectively), ABFR-KAN outperforms KD-Transformer on average by 7.94\% on the NYU site, 6.657\% on the UM site, and 30.4\% on the KKI site, while outperforming MVS-GCN on average by 13.03\% on the NYU site, 11.81\% on the UM site, and 35.9\% on the KKI site. Across the metrics for NYU and UM, the only metric where ABFR-KAN is deemed not superior is SEN; however, it should be noted that on the NYU site, the baseline models with the highest SEN values, BrainGNN and MVS-GCN, show strong evidence of false positive predictions, as indicated by their comparatively low SPE values. Compared to these, ABFR-KAN, with an average SEN of 0.8984 and SPE of 0.7159, demonstrates much better balance in meaningfully distinguishing patients with ASD from the control. The same relationship is demonstrated between SEN and SPE for the best-performing baselines (SVM, BrainGNN, MVS-GCN) on KKI in this regard, just in reverse, where the high SPE values for these three baselines indicate universal false negative predictions for these models, which is well addressed by our ABFR-KAN approach.

\begin{table*}[t]
    \centering
    \caption{Impact of various classification backbones within the ABFR-KAN framework on data from the UM site. Mean ± stdev scores are reported by averaging the results of 5-fold cross-validation. Best and second-best results are \textbf{bolded} and \underline{underlined}, respectively.}
    \vspace{0.5em}
    \resizebox{\linewidth}{!}{
    \begin{tabular}{lcccccc}
         \toprule
         Backbone & ACC & AUC & F1 & PRE & SEN & SPE \\
         \midrule
         EfficientNetV2 & 0.6818 ± 0.0643 & 0.6240 ± 0.0933 & \underline{0.7142 ± 0.0686} & 0.6422 ± 0.0771 & 0.8172 ± 0.1130 & \underline{0.6772 ± 0.0645} \\
         \rowcolor{green!17.5}
         ViT & \textbf{0.7182 ± 0.0530} & \textbf{0.6426 ± 0.1395} & 0.6999 ± 0.1311 & \textbf{0.7169 ± 0.1104} & 0.7397 ± 0.2197 & \textbf{0.6984 ± 0.0780} \\
         DeiT & 0.6909 ± 0.0782 & \underline{0.6242 ± 0.0934} & 0.7126 ± 0.1011 & \underline{0.6595 ± 0.0877} & \underline{0.8205 ± 0.1947} & 0.6766 ± 0.0779 \\
         Vim & \underline{0.6909 ± 0.0340} & 0.5856 ± 0.0764 & \textbf{0.7227 ± 0.0882} & 0.6309 ± 0.0505 & \textbf{0.8594 ± 0.1569} & 0.6740 ± 0.0560 \\
         \bottomrule
    \end{tabular}
    }
    \label{tab:backboneablation}
\end{table*}

\begin{table*}[t]
    \centering
    \caption{Impact of various anchor selection and patch sampling methods on the UM site using the baseline MLP-only method. Mean±stdev scores are reported by averaging the results of 5-fold cross-validation. Best and second-best results are \textbf{bolded} and \underline{underlined}, respectively.}
    \vspace{0.5em}
    \Huge{
    \resizebox{\linewidth}{!}{
    \begin{tabular}{llcccccc}
         \toprule
         Anchor Selection & Patch Sampling & ACC & AUC & F1 & PRE & SEN & SPE  \\
         \midrule
         Grid-based & Random & \underline{0.7000 ± 0.0545} & \underline{0.6347 ± 0.0840} & \textbf{0.7348 ± 0.0290} & \underline{0.6609 ± 0.0353} & \textbf{0.8395 ± 0.1040} & \underline{0.6899 ± 0.0871} \\
         Random & Random & 0.6636 ± 0.0364 & 0.5639 ± 0.0962 & \underline{0.7030 ± 0.0752} & 0.6124 ± 0.0545 & \underline{0.8359 ± 0.1410} & 0.6427 ± 0.0400 \\
         Grid-based & Multi-scale & 0.7000 ± 0.0617 & 0.6342 ± 0.1051 & 0.6226 ± 0.3131 & 0.5349 ± 0.2689 & 0.7470 ± 0.3784 & 0.6674 ± 0.0967 \\
         \rowcolor{green!17.5}
         Random & Multi-scale & \textbf{0.7182 ± 0.0530} & \textbf{0.6426 ± 0.1395} & 0.6999 ± 0.1311 & \textbf{0.7169 ± 0.1104} & 0.7397 ± 0.2197 & \textbf{0.6984 ± 0.0780} \\
         \bottomrule
    \end{tabular}
    }}
    \label{tab:archablation}
\end{table*}

\begin{table*}[t]
    \centering
    \caption{Impact of various KAN variants within the ABFR-KAN framework. Mean±stdev scores are reported by averaging the scores of 5-fold cross-validation. Best and second-best results are \textbf{bolded} and \underline{underlined}, respectively. ($\dagger$ indicates  KAN variants that improved upon the ACC of the baseline in a statistically significant manner, and ($\ddagger$) denotes the KAN variant that most improved upon the ACC of the baseline in a statistically significant manner.}
    \vspace{0.5em}
    \resizebox{\linewidth}{!}{
    \begin{tabular}{lcccccc}
         \toprule
         \rowcolor{gray!25}
         \multicolumn{7}{c}{\textbf{\large NYU Site Classification Performance}} \\
         \midrule
         KAN Variant & ACC & AUC & F1 & PRE & SEN & SPE  \\
         \midrule
         Baseline (MLP) & 0.7079 ± 0.0388 & 0.6544 ± 0.0583 & 0.7734 ± 0.0379 & 0.6986 ± 0.0435 & 0.8784 ± 0.1024 & 0.6783 ± 0.0486 \\
         Efficient-KAN & 0.7079 ± 0.0341 & 0.6286 ± 0.0826 & 0.7701 ± 0.0525 & 0.7022 ± 0.0490 & 0.8795 ± 0.1501 & 0.6793 ± 0.0352 \\
         \rowcolor{green!17.5}
         FastKAN ($\ddagger$) & \textbf{0.7427 ± 0.0342} & \textbf{0.7041 ± 0.0708} & \textbf{0.8003 ±  0.0200} & \underline{0.7267 ± 0.0469} & \underline{0.8984 ± 0.0646} & \textbf{0.7159 ± 0.0434} \\
         FasterKAN ($\dagger$) & \underline{0.7259 ± 0.0716} & 0.6602 ± 0.0359 & \underline{0.7954 ± 0.0429} & 0.7039 ± 0.0629 & \textbf{0.9184 ± 0.0248} & \underline{0.6925 ± 0.0838} \\
         Wav-KAN & 0.6842 ± 0.0218 & \underline{0.6672 ± 0.0331} & 0.7078 ± 0.0290 & \textbf{0.7585 ± 0.0524} & 0.6732 ± 0.0759 & 0.6842 ± 0.0311 \\
         ChebyKAN & 0.6904 ± 0.0447 & 0.6611 ± 0.0493 & 0.7315 ± 0.0615 & 0.7204 ± 0.0350 & 0.7558 ± 0.1306 & 0.6793 ± 0.0363 \\
         \midrule
         \rowcolor{gray!25}
         \multicolumn{7}{c}{\textbf{\large UM Site Classification Performance}} \\
         \midrule
         KAN Variant & ACC & AUC & F1 & PRE & SEN & SPE  \\
         \midrule
         Baseline (MLP) & 0.7182 ± 0.0530 & 0.6426 ± 0.1395 & 0.6999 ± 0.1311 & \underline{0.7169 ± 0.1104} & 0.7397 ± 0.2197 & 0.6984 ± 0.0780 \\
         Efficient-KAN & 0.7091 ± 0.0464 & \underline{0.6769 ± 0.1095} & 0.7108 ± 0.1075 & 0.6715 ± 0.0684 & 0.7688 ± 0.1637 & 0.6956 ± 0.0626 \\
         FastKAN & 0.7091 ± 0.0464 & 0.6471 ± 0.0287  & 0.7123 ± 0.0926  & 0.6683 ± 0.0482 & 0.7745 ± 0.1641 & 0.6845 ± 0.0381  \\
         FasterKAN & 0.7091 ± 0.0464 & 0.6203 ± 0.0490 & 0.7093 ± 0.1027 & 0.6813 ± 0.0761 & \underline{0.7859 ± 0.2114} & 0.6777 ± 0.0278 \\
         \rowcolor{green!17.5}
         Wav-KAN ($\ddagger$) & \textbf{0.7727 ± 0.0287} & \textbf{0.7184 ± 0.0620} & \textbf{0.7682 ± 0.0482} & \textbf{0.7840 ± 0.1086} & 0.7837 ± 0.1314 & \textbf{0.7768 ± 0.0270} \\
         ChebyKAN ($\dagger$) & \underline{0.7273 ± 0.0498} & 0.6345 ± 0.1052 & \underline{0.7443 ± 0.0838} & 0.6801 ± 0.0374 & \textbf{0.8412 ± 0.1742} & \underline{0.7061 ± 0.0794} \\
         \midrule
        \rowcolor{gray!25}
        \multicolumn{7}{c}{\textbf{\large KKI Site Classification Performance}} \\
        \midrule
        KAN Variant & ACC & AUC & F1 & PRE & SEN & SPE \\
        \midrule
        Baseline (MLP) & \underline{0.7455 ± 0.0600} & \underline{0.6248 ± 0.1552} & \underline{0.6457 ± 0.1855} & 0.7076 ± 0.1774 & \underline{0.6200 ± 0.2159} & \underline{0.7148 ± 0.1162} \\
        Efficient-KAN & 0.6727 ± 0.0445 & 0.5110 ± 0.1895 & 0.3455 ± 0.3007 & 0.4533 ± 0.3942 & 0.3300 ± 0.3219 & 0.5983 ± 0.0889 \\
        FastKAN & 0.6909 ± 0.0445 & 0.5143 ± 0.2367 & 0.3931 ± 0.3248 & 0.4167 ± 0.3416 & 0.3800 ± 0.3250 & 0.6257 ± 0.1042 \\
        FasterKAN & 0.7091 ± 0.0680 & 0.5910 ± 0.1871 & 0.5400 ± 0.1718 & \underline{0.7833 ± 0.1944} & 0.5100 ± 0.2871 & 0.6812 ± 0.0974 \\
        \rowcolor{green!17.5}
        Wav-KAN $\ddagger$ & \textbf{0.7636 ± 0.0445} & \textbf{0.6505 ± 0.0869} & \textbf{0.6588 ± 0.1329} & \textbf{0.8033 ± 0.1675} & \textbf{0.6300 ± 0.2015} & 0.7388 ± 0.0648 \\
        ChebyKAN & 0.6000 ± 0.0445 & 0.4744 ± 0.0887 & 0.0000 ± 0.0000 & 0.0000 ± 0.0000 & 0.0000 ± 0.0000 &  0.5000 ± 0.0000 \\
         \bottomrule
    \end{tabular}
    }
    \label{tab:kanablation}
\end{table*}

\textit{Statistical Analysis:} To validate the statistical significance of our reported improvements over baselines, we utilized one-tailed paired t-tests over the reported performance metrics. On the NYU site, ABFR-KAN demonstrates statistically significant improvement over KD-Transformer in terms of ACC, F1 (both $p < 0.01$), and SPE ($p < 0.05$) and over MVS-GCN on ACC, AUC, PRE (all $p < 0.05$), and SPE ($p < 0.01$). On the UM site, ABFR-KAN demonstrates statistically significant improvement over RandomFR in terms of ACC ($p < 0.01$) and SPE ($p < 0.05$), over KD-Transformer on ACC, F1, PRE, and SPE (all $p < 0.01$), and over MVS-GCN in terms of ACC, AUC (both $p < 0.01$), PRE, and SPE (both $p < 0.05$). This trend of ABFR-KAN offering statistically significant improvements continues for the poorly performing baselines across both ABIDE I sites. On the KKI site from ABIDE II, ABFR-KAN demonstrated statistically significant improvements on all metrics but SPE for the non-RandomFR baselines.

\subsubsection{Cross-Site Generalizability}

Table~\ref{tab:crosssiteresults} shows the cross-site generalizability of ABFR-KAN compared to baseline methods. We explore cross-site generalizability in two directions: first, we examine how well models trained on the NYU site generalize when tested on data from the UM site; then we explore the opposite, where models trained on UM are tested on NYU. From the results, it is evident that despite an expected drop in performance compared to the in-domain cross-validation experiments, ABFR-KAN is the superior model in terms of generalizability across both settings compared to the other baselines, albeit they are more competitive in this setting than in the single-site experiments. 

In both settings, ABFR-KAN achieves the strongest performance of the compared models on three of the six metrics. In the NYU $\rightarrow$ UM setup, ABFR-KAN achieves the highest values across AUC, F1, and PRE. On UM $\rightarrow$ NYU, ABFR-KAN achieves the best performance on ACC, AUC, and PRE. As with the discussion of the single-site performance of the models, a caveat needs to be added when viewing the results of both GCN and BrainGNN on the NYU $\rightarrow$ UM experiment, as well as BrainNetCNN and GCN on UM $\rightarrow$ NYU. In both instances, the relationship between the reported SEN and SPE values for these models indicates issues in terms of clinical viability, as there is a high indication of both false positives (BrainGNN on NYU $\rightarrow$ UM, GCN on UM$\rightarrow$ NYU) and missed positives (GCN on NYU $\rightarrow$ UM, BrainNetCNN on UM$\rightarrow$ NYU). ABFR-KAN, in comparison, achieves a much better balance between R and SPE in both settings, indicating better clinical viability.

However, it should be noted that while ABFR-KAN achieved the strongest AUC in the NYU → UM cross-site setting (0.7181), the corresponding classification accuracy remained relatively low at 0.5517, which was below the RandomFR baseline (0.6109). This discrepancy between AUC and ACC indicates that the model retains meaningful discriminative capability in terms of ranking ASD versus control subjects, but that the learned decision boundary does not transfer robustly across acquisition sites. Since AUC evaluates separability across all possible thresholds whereas ACC depends on a fixed threshold of 0.5, these results suggest that ABFR-KAN produces informative probability estimates under domain shift, but that calibration deteriorates when evaluated on unseen sites.

This behavior, along with the general asymmetry of the cross-site results, with models trained on NYU generalizing differently to UM compared to the reverse direction, is to be expected due to the disparate data distributions between the two sites. As described in Section~\ref{subsec:data}, the NYU and UM cohorts differ in multiple respects, including age ranges, sex distributions, cohort balance, and scanner acquisition characteristics. In particular, the NYU cohort contains a broader age range and a more imbalanced class distribution, whereas the UM cohort is more tightly age-constrained and balanced. The domain shift caused by these differences means that the properties learned from one site may not be equally transferable to the other.

\textit{Statistical Analysis:} Using the same one-tailed paired t-test as before, we investigate the statistical significance of our proposed ABFR-KAN framework over existing methods. On the NYU $\rightarrow$ UM setup, ABFR-KAN achieves a statistically significant improvement over RandomFR in AUC, F1, and PRE (all $p < 0.01$). On UM $\rightarrow$ NYU, ABFR-KAN improves in a statistically significant manner over GCN in terms of AUC ($p < 0.05$) and SPE ($p < 0.01$), over GAT on ACC, PRE (both $p < 0.05$), AUC, and SPE (both $p < 0.01$), and over BrainGNN in ACC, AUC, PRE (all $p < 0.05$), and F1 ($p < 0.01$). Like before, the trend of ABFR-KAN offering statistically significant improvements continues across lesser-performing baselines in both setups.

\subsubsection{Comparison to Atlas-Based Techniques}

In order to assess the performance of ABFR-KAN against ``gold-standard'' atlas-based techniques, Table~\ref{tab:atlasresults} shows a direct comparison against several such techniques. The compared-against methods are:~\cite{spera2019evaluation}, whose method utilizes the Harvard-Oxford anatomical atlas;~\cite{elnakieb2021identifying}, which uses the Johns Hopkins University white matter atlas; and ~\cite{yang2024identification}, which uses the Brodmann atlas. The results demonstrate that ABFR-KAN is capable of performing comparably to, or even outperforming the atlas-based techniques on certain metrics.

\begin{table}[t]
    \centering
    \caption{Performance of ABFR-KAN against several atlas-based techniques for ASD diagnosis on the NYU site. Best and second-best results are \textbf{bolded} and \underline{underlined}, respectively.}
    \footnotesize
    \begin{tabular}{lcccc}
         \toprule
         Model & ACC & AUC & SEN & SPE \\
         \midrule
         \cite{spera2019evaluation} & 0.6300 & \textbf{0.7500} & 0.4800 & \textbf{0.8300} \\
         \cite{elnakieb2021identifying} & 0.6978 & \textbf{---} & \underline{0.7200} & 0.6700 \\
         \cite{yang2024identification} & \underline{0.7294} & \textbf{---} & 0.7083 & \underline{0.7470} \\
         \midrule
         \rowcolor{green!17.5}
         Ours & \textbf{0.7427} & \underline{0.7041} & \textbf{0.8984} & 0.7159 \\
         \bottomrule
    \end{tabular}
    \label{tab:atlasresults}
\end{table}

\subsubsection{Ablation Experiments}
\label{subsubsec:ablation}
In order to justify the existence of each of our proposed components within our ABFR-KAN framework, as well as to determine the best-performing KAN variants and configurations, we conduct a series of ablation experiments. Tables~\ref{tab:backboneablation} and~\ref{tab:archablation} show the impact of various classification backbones and anchor selection/patch sampling configurations within the ABFR-KAN framework, respectively. These experiments were conducted on the UM site of ABIDE I, as the same experiments on the NYU site were previously conducted and thoroughly analyzed in our previous work \cite{ward2025improving}.

From the results shown in Table~\ref{tab:backboneablation}, it is clear that the use of ViT as the classification backbone offers the best performance, with it achieving top performance across four of the six methods. This validates our previous findings~\cite{ward2025improving}, where we determined that ViT was a better choice than DeiT. The Mamba-based method Vim achieves the best performance on the F1 and R metrics, but overall, it performs worse than ViT. The CNN-based method, EfficientNetV2, performs the worst overall, which aligns with the established knowledge that transformer-based methods often outperform CNN-based ones~\cite{dosovitskiy2020image}.

\begin{figure}[t]
\centering
\resizebox{0.85\linewidth}{!}{%
\begin{tikzpicture}
\begin{axis}[
    xlabel={Number of scales},
    ylabel={FC variance},
    xtick={1,2,3},
    ytick={0.0175,0.0200,0.0225,0.0250,0.0275,0.0300,0.0325,0.0350},
    ymin=0.016,
    ymax=0.036,
    grid=both,
]

\addplot[
    mark=*,
    mark size=2.5pt,
]
coordinates {
    (1, 0.03432292)
    (2, 0.02056173)
    (3, 0.01647731)
};

\end{axis}
\end{tikzpicture}%
}
\caption{Impact of differing scale values in our proposed multi-scale patch sampling method on reducing sampling variance in FC estimation. Increasing the number of iterations beyond single-pass sampling (the baseline) demonstrably reduces sample variance, which offers benefits that include improved feature stability, a reduction in the risk of overfitting to sampling artifacts, and improved reproducibility between sites.}
\label{fig:iterationablation}
\end{figure}

Table~\ref{tab:archablation} demonstrates that our proposed architectural configuration involving the use of random anchor selection followed by multi-scale patch sampling is superior compared to the baseline grid-based anchor selection approach followed by random patch sampling. Again, this validates our findings in our previous work \cite{ward2025improving}. Interestingly, the results also show that simply applying randomized anchor selection while keeping the random patch sampling can perform worse than the original baseline configuration that uses neither proposed component. We hypothesize that this occurs because randomized anchor selection increases inter-subject variability and representational diversity, which, without accompanying multi-scale sampling, may lead to less stable FC estimation and reduced robustness due to the additional stochasticity introduced by randomized anchors. As for why the use of grid-based anchors and multi-scale patch sampling also exhibits worse performance compared to the baseline, we posit that grid-based anchors, while structurally biased, may provide a spatially consistent reference structure that can partially compensate for the limitations of single-pass patch sampling.

We briefly mentioned this in Section~\ref{subsec:iterative}, but would now like to draw attention to Fig.~\ref{fig:iterationablation}, which both demonstrates that increasing the number of iterations reduces the FC sampling variance, which has numerous benefits, and serves as an ablation experiment on the number of iterations themselves. Table~\ref{tab:kanablation} similarly investigates the relative behavior of various KAN variants when incorporated into the ABFR-KAN framework, with the primary goal being comparative analysis rather than formal hyperparameter or architecture optimization. A detailed analysis of the computational complexity and efficiency of each variation is provided in Section~\ref{subsec:discussion}, based on Figure~\ref{fig:performance_tradeoff}. For brevity, we report only the results of the best-performing configuration involving these KAN variants in this table. Additional experiments that showcase the full range of possible configurations for each of these variants on both ABIDE I sites are reported in Appendix~\ref{apdx:additionalexperiments}.

From our experiments, we found that the KAN variants perform differently across the three sites. This is, perhaps, to be expected given the diversity in patient populations between the two sites (See Section~\ref{subsec:data}) and varying acquisition parameters between them. On the NYU site, the FastKAN variant in a KAN-KAN configuration (where FastKAN blocks replaced the MLP components in both the ViT encoder and the classification head) came out on top, outperforming the second-best method (FasterKAN, also in a KAN-KAN configuration) by 1.41\%, on average. On the UM site, Wav-KAN (in an MLP-KAN configuration where only the MLP classification head was replaced by Wav-KAN) was clearly superior to the other KAN variants, achieving a 4.51\% improvement on average over the second-best variant, ChebyKAN. On KKI, Wav-KAN again demonstrated supremacy, outperforming the second-best FasterKAN by 7.18\% on average.

\begin{figure*}[t]
\centering

\begin{tikzpicture}

% ===================== LEFT PLOT =====================
\begin{axis}[
    width=0.46\textwidth,
    height=7cm,
    xlabel={Number of Parameters},
    ylabel={Accuracy},
    xmin=90000,
    xmax=210000,
    ymin=0.55,
    ymax=0.76,
    grid=both,
    mark size=3,
    tick label style={font=\small},
    label style={font=\small},
    legend to name=combinedlegend,
    legend columns=4,
    legend style={
        font=\scriptsize,
        draw=none,
        /tikz/every even column/.append style={column sep=0.4cm}
    }
]

% ---------------- DATA ----------------
\addplot[only marks, mark=*, color=blue]
coordinates {(100066,0.7079)};
\addlegendentry{MLP-MLP}

\addplot[only marks, mark=square*, color=red]
coordinates {(116032,0.7079)};
\addlegendentry{MLP-EfficientKAN}

\addplot[only marks, mark=triangle*, color=green!70!black]
coordinates {(116306,0.7309)};
\addlegendentry{MLP-FastKAN}

\addplot[only marks, mark=diamond*, color=orange]
coordinates {(114464,0.6903)};
\addlegendentry{MLP-FasterKAN}

\addplot[only marks, mark=pentagon*, color=purple]
coordinates {(106948,0.6842)};
\addlegendentry{MLP-WavKAN}

\addplot[only marks, mark=star, color=brown]
coordinates {(108656,0.6845)};
\addlegendentry{MLP-ChebyKAN}

\addplot[only marks, mark=*, color=cyan]
coordinates {(185698,0.7022)};
\addlegendentry{EfficientKAN-MLP}

\addplot[only marks, mark=square*, color=magenta]
coordinates {(186850,0.7195)};
\addlegendentry{FastKAN-MLP}

\addplot[only marks, mark=triangle*, color=teal]
coordinates {(175714,0.7197)};
\addlegendentry{FasterKAN-MLP}

\addplot[only marks, mark=diamond*, color=olive]
coordinates {(132706,0.6844)};
\addlegendentry{WavKAN-MLP}

\addplot[only marks, mark=pentagon*, color=gray]
coordinates {(143346,0.6904)};
\addlegendentry{ChebyKAN-MLP}

\addplot[only marks, mark=*, color=black]
coordinates {(201664,0.6724)};
\addlegendentry{EfficientKAN-EfficientKAN}

\addplot[only marks, mark=square*, color=red!70!black]
coordinates {(203090,0.7427)};
\addlegendentry{FastKAN-FastKAN}

\addplot[only marks, mark=triangle*, color=blue!70!black]
coordinates {(190112,0.7259)};
\addlegendentry{FasterKAN-FasterKAN}

\addplot[only marks, mark=diamond*, color=green!50!black]
coordinates {(139588,0.6726)};
\addlegendentry{WavKAN-WavKAN}

\addplot[only marks, mark=star, color=orange!80!black]
coordinates {(152048,0.5731)};
\addlegendentry{ChebyKAN-ChebyKAN}

\end{axis}

% ===================== RIGHT PLOT =====================
\begin{axis}[
    width=0.46\textwidth,
    height=7cm,
    xlabel={Training Time (s)},
    ylabel={Accuracy},
    xmin=35,
    xmax=120,
    ymin=0.55,
    ymax=0.76,
    grid=both,
    mark size=3,
    tick label style={font=\small},
    label style={font=\small},
    at={(0.52\textwidth,0)},
    anchor=south west
]

% ---------------- SAME DATA ----------------
\addplot[only marks, mark=*, color=blue]
coordinates {(41.46,0.7079)};

\addplot[only marks, mark=square*, color=red]
coordinates {(54.34,0.7079)};

\addplot[only marks, mark=triangle*, color=green!70!black]
coordinates {(46.88,0.7309)};

\addplot[only marks, mark=diamond*, color=orange]
coordinates {(42.48,0.6903)};

\addplot[only marks, mark=pentagon*, color=purple]
coordinates {(46.98,0.6842)};

\addplot[only marks, mark=star, color=brown]
coordinates {(43.64,0.6845)};

\addplot[only marks, mark=*, color=cyan]
coordinates {(94.30,0.7022)};

\addplot[only marks, mark=square*, color=magenta]
coordinates {(55.78,0.7195)};

\addplot[only marks, mark=triangle*, color=teal]
coordinates {(49.87,0.7197)};

\addplot[only marks, mark=diamond*, color=olive]
coordinates {(113.15,0.6844)};

\addplot[only marks, mark=pentagon*, color=gray]
coordinates {(52.37,0.6904)};

\addplot[only marks, mark=*, color=black]
coordinates {(100.17,0.6724)};

\addplot[only marks, mark=square*, color=red!70!black]
coordinates {(55.91,0.7427)};

\addplot[only marks, mark=triangle*, color=blue!70!black]
coordinates {(52.73,0.7259)};

\addplot[only marks, mark=diamond*, color=green!50!black]
coordinates {(115.03,0.6726)};

\addplot[only marks, mark=star, color=orange!80!black]
coordinates {(51.86,0.5731)};

\end{axis}

\end{tikzpicture}

\vspace{0.5cm}

\pgfplotslegendfromname{combinedlegend}

\caption{Trade-off analysis between classification performance on the NYU site, model complexity, and computational cost across different KAN architectural configurations.
Left: Accuracy versus number of trainable parameters.
Right: Accuracy versus total training time across all five folds of 5-fold cross-validation.}
\label{fig:performance_tradeoff}

\end{figure*}

\subsection{Discussion}
\label{subsec:discussion}
\noindent\textit{Reducing the Reliance on Atlases:} A core contribution of our work is the removal of reliance on predefined atlases through randomized anchor selection. Unlike the baseline grid-based approach, which imposes a rigid spatial arrangement, our method randomly selects anchor patches from valid GM regions, resulting in improved anatomical conformity and reduced structural bias, as demonstrated in Figs.~\ref{fig:anchorhistogram} and ~\ref{fig:anchorpatchcenters}. Importantly, these anchors are sampled once at the dataset level and are subsequently fixed and shared across all subjects, providing a common reference space for functional connectivity computation. Thus, the individual-specificity of the resulting representations does not arise from anchor selection itself, but from the independently sampled local brain patches used to construct each subject’s functional representation.

The use of randomized rather than grid-based anchors may nevertheless influence downstream performance by altering the spatial reference against which subject-specific patches are characterized. Our ablation experiments suggest that randomized anchor selection is most effective when paired with multi-scale patch sampling. In isolation, randomized anchors can introduce additional variability into the functional representation and may reduce stability, whereas multi-scale sampling helps mitigate this effect by aggregating multiple stochastic estimates of functional connectivity. This interaction may explain why the full combination of randomized anchor selection and multi-scale patch sampling performs better than configurations using either component alone. Overall, randomized anchor selection should therefore be viewed primarily as a mechanism for reducing atlas-related structural bias and improving anatomical conformity, while subject-level patch sampling provides the source of individual-specific functional variability.

\vspace{0.157em}

\noindent\textit{Complexity and Efficiency of the Explored KAN Variants:} Fig.~\ref{fig:performance_tradeoff} shows the trade-off in training time and number of parameters for the five different KAN variants we tested in this work (Efficient-KAN, FastKAN, FasterKAN, Wav-KAN, ChebyKAN). We show this trade-off across each of the three architectural configurations we investigated (MLP-KAN, KAN-MLP, KAN-KAN), and the baseline (MLP-MLP). As expected, the baseline method using only MLPs is both the fastest to train (41.46 seconds in total), and uses the smallest number of parameters (100,066). This is consistent with known characteristics of KANs, with them often increasing parameter counts with the trade-off of improved performance \cite{wang2024expressiveness}.

Despite the increase in model complexity and decrease in efficiency, we argue that the performance trade-off is well justified by the results reported in Tables~\ref{tab:singlesiteresults} and~\ref{tab:crosssiteresults}, especially. Take, for instance, Wav-KAN in an MLP-KAN configuration (denoted by the red pentagon in Fig.~\ref{fig:performance_tradeoff}). This configuration offered the best performance on the single-site UM experiment, improving on average over the baseline method by 6.56\%, while adding only 6,882 parameters and 5.52 seconds to the total training time. However, if training complexity and efficiency are top priorities, practitioners must carefully weigh whether KAN integration is worth the extra overhead.

\vspace{0.157em}

\noindent\textit{The Role of KANs:} Replacing traditional MLP components with KAN variants plays a key role in the performance gains offered by ABFR-KAN. KANs offer increased expressiveness through learnable univariate functions while reducing over-parameterization relative to dense MLPs operating on high-dimensional FC features. The ablation studies demonstrate that KAN-based networks consistently outperform MLP-only baselines, with performance boosts depending on both the KAN variant and its placement within the architecture. Interestingly, no single KAN variant dominates across all experimental settings. FastKAN performs best on the NYU site, while Wav-KAN achieves superior results on the UM site, suggesting that the optimal basis function may be data- and domain-dependent. This observation aligns with recent findings that KAN design choices remain task-specific~\cite{becerra2025mathematical}, and it motivates future work on adaptive or hybrid KAN configurations for neuroimaging applications.

\vspace{0.157em}

\noindent\textit{Clinical Implications:} Cross-site experiments reveal that ABFR-KAN generalizes better than existing methods when trained and tested across different imaging sites. From a clinical perspective, the improved balance between sensitivity and specificity is particularly important. Several baseline methods achieve high SEN at the cost of excessive false positives, limiting their practical utility. ABFR-KAN avoids this issue, indicating a stronger potential for real-world diagnostic support.

\vspace{0.157em}

\noindent\textit{Limitations and Future Directions:} Despite the promise of our ABFR-KAN method, there are several limitations to our work. \textit{First,} we focus exclusively on rs-fMRI data from the ABIDE repository. While cross-site evaluation partially demonstrates the generalizability of our approach, in the future, this should be extended to validate additional datasets, disorders, and imaging modalities. \textit{Second,} compared to the prevalence of ASD in the real-world population, the datasets used in this research are fairly balanced, leaving questions about how well the model would perform on a real population dataset. A potential solution to dealing with real, imbalanced datasets lies in the use of loss functions designed to handle class imbalance, such as Focal loss~\cite{lin2017focal}. \textit{Third,} although KANs offer improved interpretability in principle, we don't explicitly analyze the learned activation functions. Investigating whether these functions correspond to neurobiologically meaningful patterns would be an interesting avenue for future research. \textit{Finally,} the KAN variant and configuration selection were performed empirically on a per-site basis. Although this enabled exploration of the relative effectiveness of different KAN formulations under varying data characteristics, the selection process was not fully isolated from the evaluation protocol. As a result, the reported performance estimates may partially reflect site-specific optimization and could therefore overestimate generalization performance. Future work should employ a strictly nested cross-validation framework, where architecture and hyperparameter selection are performed exclusively within inner validation folds while keeping the outer test folds completely unseen, in order to eliminate potential information leakage and provide a more rigorous estimate of model generalizability.

\section{Conclusions}
In this work, we introduced ABFR-KAN, a novel framework for functional brain analysis that addresses the limitations of atlas-based parcellations in defining ROIs. By combining randomized, anatomy-aware anchor selection with multi-scale patch sampling, our approach reduces structural bias, improves GM coverage, and yields more robust and individualized functional representations. To make this approach useful for downstream diagnosis, we follow this data pipeline with a ViT-style classification network. In this network, we demonstrate that replacing traditional MLP components in the encoder and classification head with KANs leads to consistent and statistically significant performance gains for ASD diagnosis.

Extensive experiments on the ABIDE I and II datasets show that ABFR-KAN outperforms strong SOTA baselines in both single-site and cross-site evaluations. Ablation studies further confirm the necessity of combining randomized anchor selection with multi-scale patch sampling and KAN-based modeling, while also highlighting the importance of selecting the optimal classification backbone and KAN variant/configuration for the task at hand. Future work will explore extending ABFR-KAN to additional neurological and psychiatric disorders, as well as expanding experimentation to different datasets.

%\clearpage
\ethics{The work follows appropriate ethical standards in conducting research and writing the manuscript, following all applicable laws and regulations regarding the treatment of animals or human subjects.}

\coi{We declare that we have no conflicts of interest.}

% Data availability
\data{For this work, we trained/evaluated our models on publicly available rs-fMRI data from the ABIDE Preprocessed repository \cite{craddock2013neuro}. These data can be downloaded following the instructions at  \href{http://preprocessed-connectomes-project.org/abide/download.html}{http://preprocessed-connectomes-project.org/abide/download.html}. The code, along with pre-trained checkpoints and data samples, is available in our public GitHub repository, linked at the end of our abstract.}

\bibliography{sample}

\clearpage

\appendix

\twocolumn[{
\section{\centering Multi-Scale Patch Sampling Visualization}

\begin{center}

\begin{tabular}{c c c c c}
\includegraphics[width=0.18\textwidth]{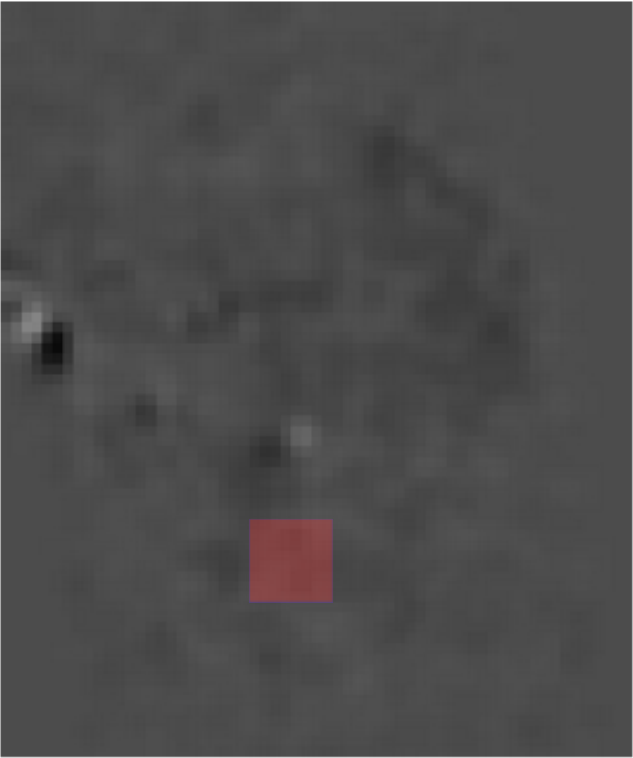}
&
\includegraphics[width=0.18\textwidth]{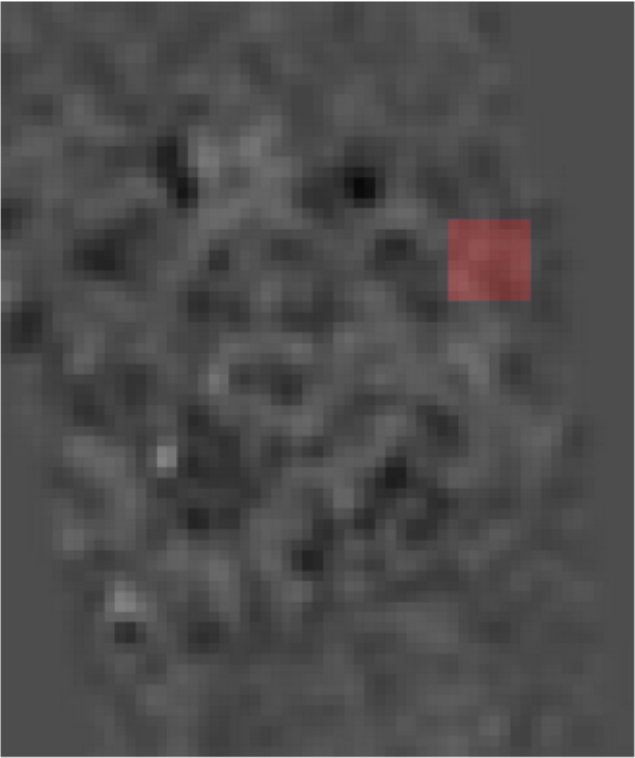}
&
\includegraphics[width=0.18\textwidth]{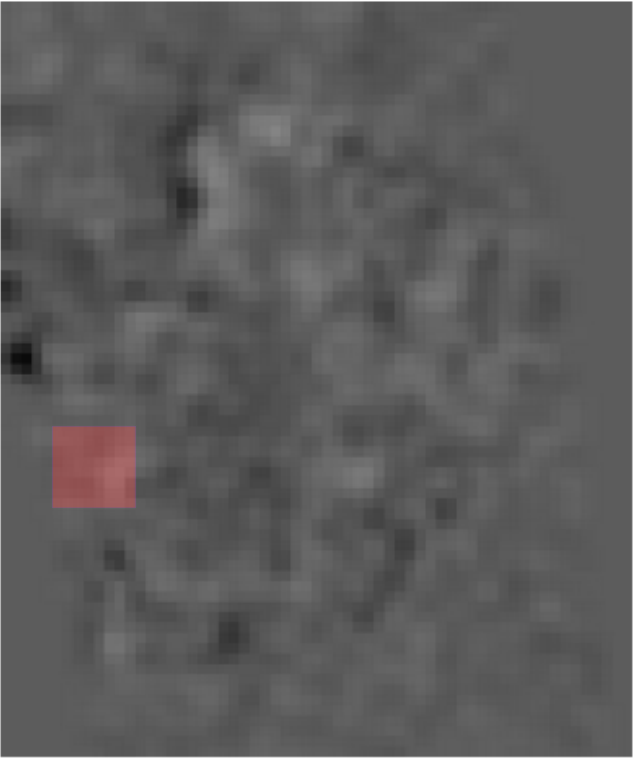}
&
\includegraphics[width=0.18\textwidth]{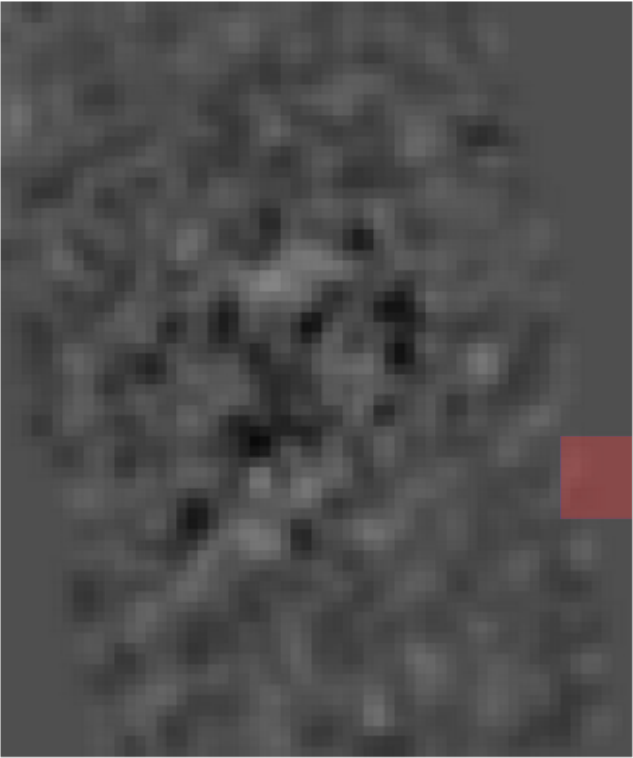}
&
\includegraphics[width=0.18\textwidth]{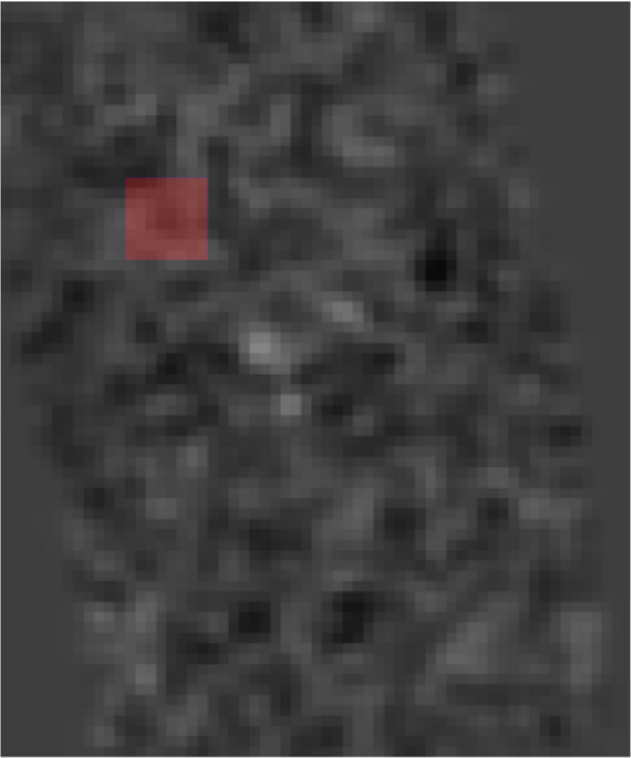}
\\

\multicolumn{5}{c}{\Large{(a) Random patch sampling}}
\\[0.75em]

\includegraphics[width=0.18\textwidth]{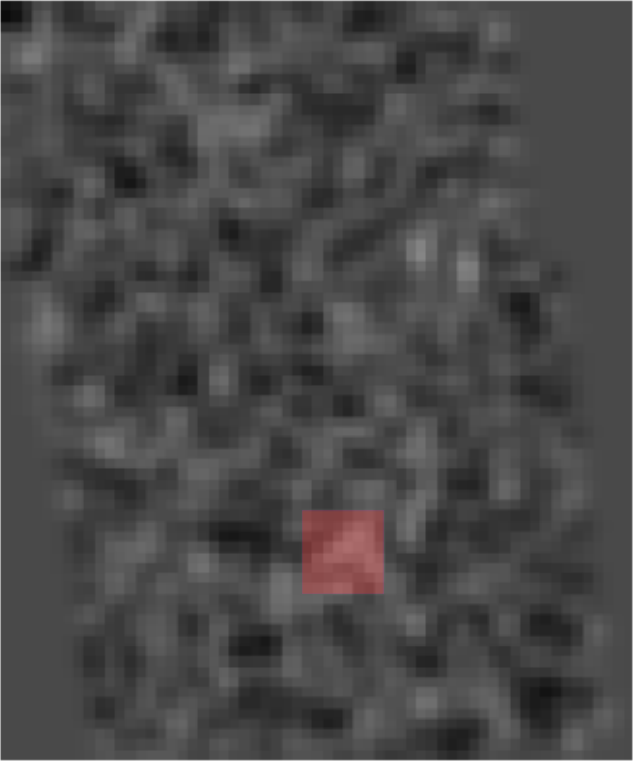}
&
\includegraphics[width=0.18\textwidth]{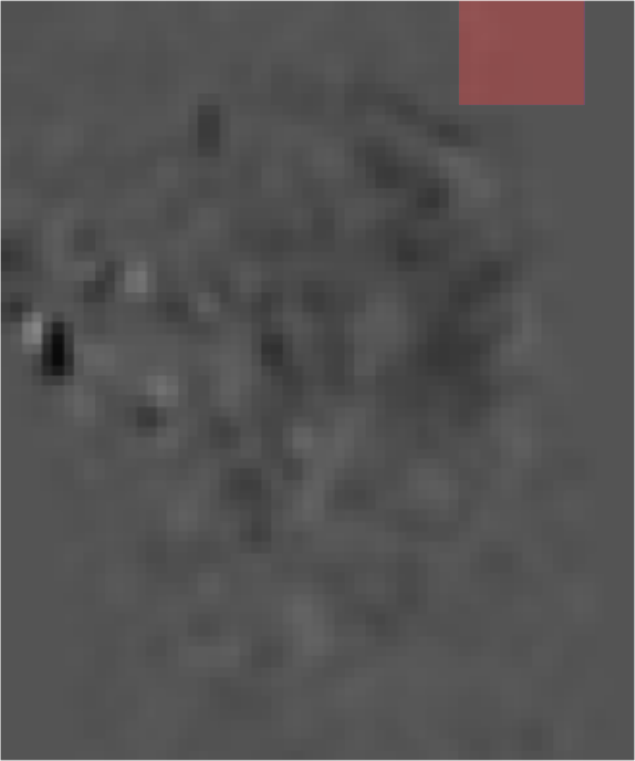}
&
\includegraphics[width=0.18\textwidth]{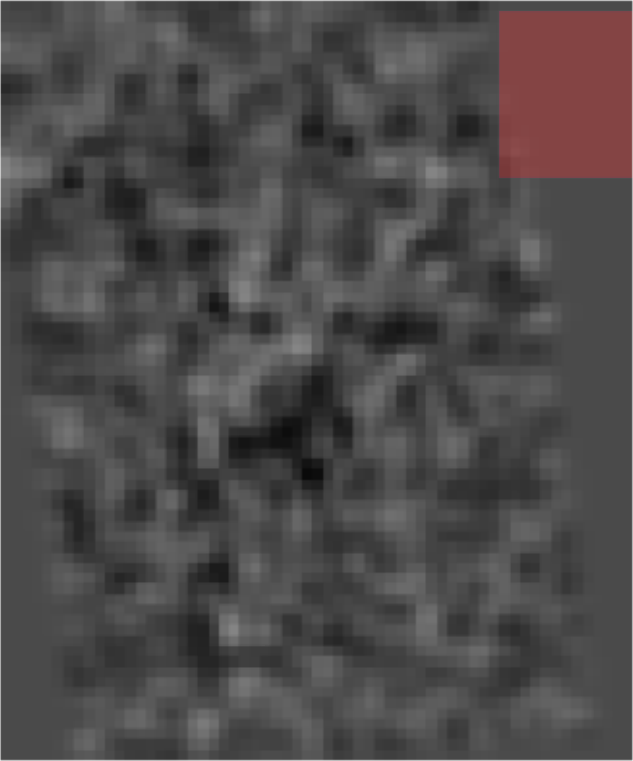}
&
\includegraphics[width=0.18\textwidth]{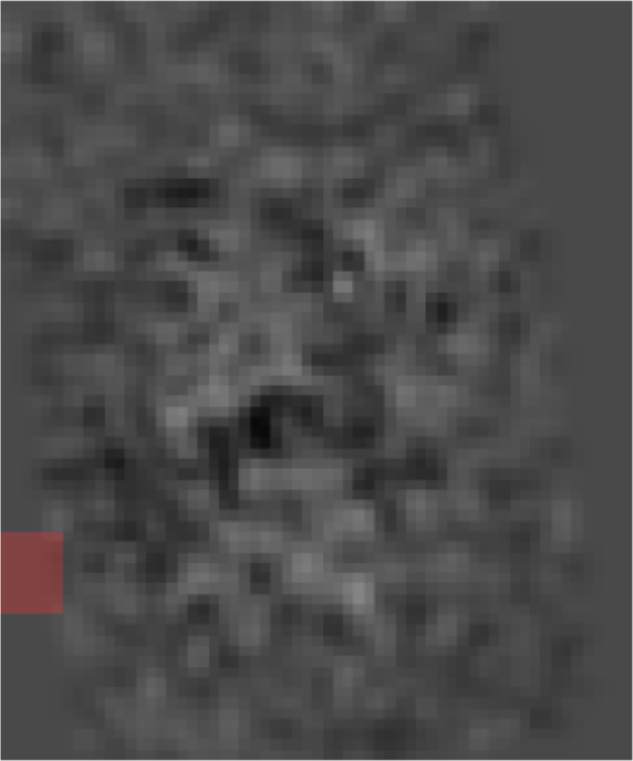}
&
\includegraphics[width=0.18\textwidth]{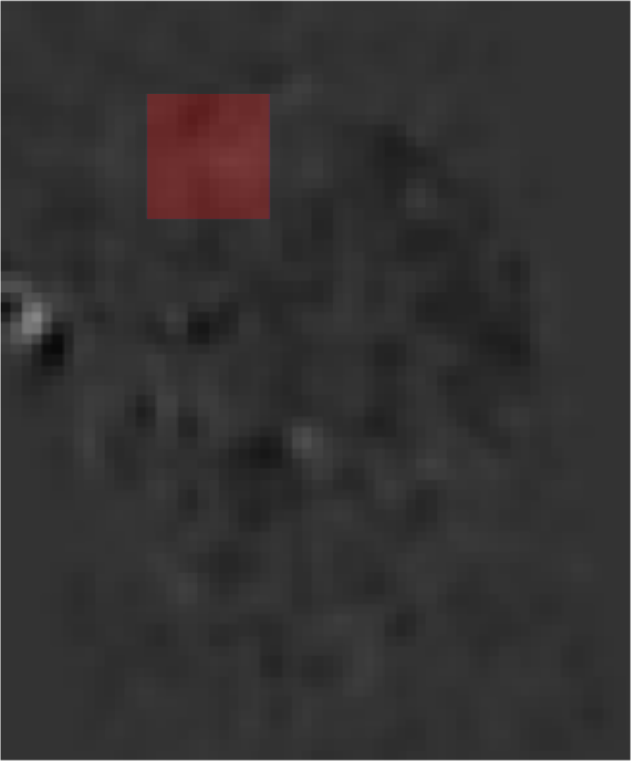}
\\

\multicolumn{5}{c}{\Large{(b) Multi-scale patch sampling}}
\end{tabular}

\captionof{figure}{
(a) The random patch sampling process. Please note how the size of the patches is consistent. 
(b) The multi-scale patch sampling process, where each subject is processed three times as a form of data augmentation, with patch sizes varying from $8 \times 8$, $12 \times 12$, and $16 \times 16$.
}

\label{fig:patchsampling}

\end{center}
}]

\clearpage

\twocolumn[
    \section{\centering Additional Experiments on KAN Variants and Configurations}
    \label{apdx:additionalexperiments}
    \captionof{table}{Impact of various KAN variants and configurations within the ABFR-KAN framework on the NYU site. Mean±stdev scores are reported by averaging the results of 5-fold cross-validation. Best and second-best results are \textbf{bolded} and \underline{underlined}, respectively.}
    \label{tab:additionalnyuresults}
    \vspace{0.5em}
    \resizebox{\linewidth}{!}{
    \begin{tabular}{lcccccc}
    \toprule
    \rowcolor{gray!25}
    \multicolumn{7}{c}{\textbf{\large Baseline}} \\
    \midrule
    Configuration & ACC & AUC & F1 & PRE & SEN & SPE \\
    \midrule
    MLP-MLP & 0.7079 ± 0.0388 & 0.6544 ± 0.0583 & 0.7734 ± 0.0379 & 0.6986 ± 0.0435 & 0.8784 ± 0.1024 & 0.6783 ± 0.0486 \\
    \midrule
    \rowcolor{gray!25}
    \multicolumn{7}{c}{\textbf{\large Efficient-KAN}} \\
    \midrule
    Configuration & ACC & AUC & F1 & PRE & SEN & SPE \\
    \midrule
    \rowcolor{green!17.5}
    MLP-KAN & \textbf{0.7079 ± 0.0341} & 0.6286 ± 0.0826 & \textbf{0.7701 ± 0.0525} & \underline{0.7022 ± 0.0490} & \textbf{0.8795 ± 0.1501} & \textbf{0.6793 ± 0.0352} \\
    KAN-MLP & \underline{0.7022 ± 0.0433} & \textbf{0.6491 ± 0.0510} & \underline{0.7665 ± 0.0516} & 0.6911 ± 0.0172 & \underline{0.8689 ± 0.1115} & \underline{0.6735 ± 0.0383} \\
    KAN-KAN & 0.6724 ± 0.0573 & \underline{0.6340 ± 0.0417} & 0.7363 ± 0.0223 & \textbf{0.7051 ± 0.0824} & 0.7958 ± 0.1146 & 0.6488 ± 0.0882 \\
    \midrule
    \rowcolor{gray!25}
    \multicolumn{7}{c}{\textbf{\large FastKAN}} \\
    \midrule
    Configuration & ACC & AUC & F1 & PRE & SEN & SPE \\
    \midrule
    MLP-KAN & \underline{0.7309 ± 0.0394} & 0.6816 ± 0.0623 & \underline{0.7807 ± 0.0303} & \textbf{0.7367 ± 0.0483} & \underline{0.8363 ± 0.0614} & \underline{0.7120 ± 0.0499} \\
    KAN-MLP & 0.7195 ± 0.0424 & \underline{0.6837 ± 0.0481} & 0.7679 ± 0.0423 & \underline{0.7328 ± 0.0504} & 0.8179 ± 0.0968 & 0.7037 ± 0.0456 \\
    \rowcolor{green!17.5}
    KAN-KAN & \textbf{0.7427 ± 0.0342} & \textbf{0.7041 ± 0.0708} & \textbf{0.8003 ± 0.0200} & 0.7267 ± 0.0469 & \textbf{0.8984 ± 0.0646} & \textbf{0.7159 ± 0.0434} \\
    \midrule
    \rowcolor{gray!25}
    \multicolumn{7}{c}{\textbf{\large FasterKAN}} \\
    \midrule
    Configuration & ACC & AUC & F1 & PRE & SEN & SPE \\
    \midrule
    MLP-KAN & 0.6903 ± 0.0380 & 0.6355 ± 0.0689 & 0.7541 ± 0.0499 & 0.6929 ± 0.0472 & 0.8479 ± 0.1300 & 0.6630 ± 0.0398 \\
    KAN-MLP & \underline{0.7197 ± 0.0364} & \underline{0.6380 ± 0.0562} & \underline{0.7735 ± 0.0546} & \textbf{0.7154 ± 0.0283} & \underline{0.8589 ± 0.1312} & \textbf{0.6961 ± 0.0272} \\
    \rowcolor{green!17.5}
    KAN-KAN & \textbf{0.7259 ± 0.0716} & \textbf{0.6602 ± 0.0359} & \textbf{0.7954 ± 0.0429} & \underline{0.7039 ± 0.0629} & \textbf{0.9184 ± 0.0248} & \underline{0.6925 ± 0.0838} \\
    \midrule
    \rowcolor{gray!25}
    \multicolumn{7}{c}{\textbf{\large Wav-KAN}} \\
    \midrule
    Configuration & ACC & AUC & F1 & PRE & SEN & SPE \\
    \midrule
    \rowcolor{green!17.5}
    MLP-KAN & \underline{0.6842 ± 0.0218} & \textbf{0.6672 ± 0.0331} & \underline{0.7078 ± 0.0290} & \textbf{0.7585 ± 0.0524} & \underline{0.6732 ± 0.0759} & \textbf{0.6842 ± 0.0311} \\
    KAN-MLP & \textbf{0.6844 ± 0.0265} & 0.6191 ± 0.0999 & \textbf{0.7349 ± 0.0352} & 0.7177 ± 0.0612 & \textbf{0.7774 ± 0.1270} & 0.6720 ± 0.0364 \\
    KAN-KAN & 0.6726 ± 0.0201 & \underline{0.6466 ± 0.0185} & 0.6933 ± 0.0542 & \underline{0.7540 ± 0.0750} & 0.6637 ± 0.1157 & \underline{0.6733 ± 0.0181} \\
    \midrule
    \rowcolor{gray!25}
    \multicolumn{7}{c}{\textbf{\large ChebyKAN}} \\
    \midrule
    Configuration & ACC & AUC & F1 & PRE & SEN & SPE \\
    \midrule
    MLP-KAN & \underline{0.6845 ± 0.0442} & \underline{0.6526 ± 0.0459} & \textbf{0.7545 ± 0.0441} & \underline{0.6803 ± 0.0407} & \underline{0.8568 ± 0.1045} & \underline{0.6546 ± 0.0408} \\
    \rowcolor{green!17.5}
    KAN-MLP & \textbf{0.6904 ± 0.0447} & \textbf{0.6611 ± 0.0493} & \underline{0.7315 ± 0.0615} & \textbf{0.7204 ± 0.0350} & 0.7558 ± 0.1306 & \textbf{0.6793 ± 0.0363} \\
    KAN-KAN & 0.5731 ± 0.0132 & 0.5054 ± 0.1340 & 0.7285 ± 0.0106 & 0.5731 ± 0.0132 & \textbf{1.0000 ± 0.0000} & 0.5000 ± 0.0000 \\
    \bottomrule
    \end{tabular}
    }
    \vspace{1em}
]
\noindent Building on the discussion of our reported results in Section~\ref{subsubsec:singlesiteresults} and Table~\ref{tab:singlesiteresults}, in this appendix, we report the full results of all of our experiments on both the NYU and UM sites to investigate the efficacy of different KAN variants and architectural configurations within our ViT-style classification network. Table~\ref{tab:additionalnyuresults} reports the results of our experimentation across five different KAN variants (Efficient-KAN, FastKAN, FasterKAN, Wav-KAN, ChebyKAN) and three different architectural configurations (MLP-KAN, KAN-MLP, KAN-KAN) on the NYU site. Table~\ref{tab:additionalumresults} reports the results of experiments with these same variants and configurations on the UM site. For the configurations, MLP-KAN refers to setups where KAN blocks replace the MLP classification head, KAN-MLP refers to setups where the MLP blocks in the transformer encoder were replaced with KAN blocks, and in the KAN-KAN setup, KANs fully replaced MLPs in the architecture.

On the NYU site, the ordering of the KAN variants, by best average performance, is: FastKAN $\rightarrow$ FasterKAN $\rightarrow$ Efficient-KAN $\rightarrow$ ChebyKAN $\rightarrow$ Wav-KAN. There is no clear winner in terms of the best architectural configuration, with two of the KAN variants (Efficient-KAN, Wav-KAN) achieving the best performance with the MLP-KAN configuration, another two (FastKAN, FasterKAN) achieving the best performance with the KAN-KAN configuration, and ChebyKAN achieving the best performance in the KAN-MLP configuration. That being said, the top two KAN variants, FastKAN and FasterKAN, use the KAN-KAN configuration, and we have previously found this configuration to perform well across thorough experimentation on the NYU site \cite{ward2025improving}. Overall, FastKAN offers a 0.63\% improvement over FasterKAN.

\begin{table*}[t]
\centering
\caption{Impact of various KAN variants and configurations within the ABFR-KAN framework on the UM site. Mean$\pm$stdev scores are reported by averaging the results of 5-fold cross-validation. Best and second-best results are \textbf{bolded} and \underline{underlined}, respectively.}
\label{tab:additionalumresults}
\resizebox{\linewidth}{!}{
\begin{tabular}{lcccccc}
\toprule
\rowcolor{gray!25}
\multicolumn{7}{c}{\textbf{\large Baseline}} \\
\midrule
Configuration & ACC & AUC & F1 & PRE & SEN & SPE \\
\midrule
MLP-MLP & 0.7182 ± 0.0530 & 0.6426 ± 0.1395 & 0.6999 ± 0.1311 & 0.7169 ± 0.1104 & 0.7397 ± 0.2197 & 0.6984 ± 0.0780 \\
\midrule
\rowcolor{gray!25}
\multicolumn{7}{c}{\textbf{\large Efficient-KAN}} \\
\midrule
Configuration & ACC & AUC & F1 & PRE & SEN & SPE \\
\midrule
MLP-KAN & \underline{0.6818 ± 0.0407} & 0.6250 ± 0.0754 & 0.6512 ± 0.1252 & \textbf{0.7325 ± 0.1444} & 0.6751 ± 0.2356 & 0.6615 ± 0.0507 \\
KAN-MLP & \textbf{0.7091 ± 0.0464} & \underline{0.6377 ± 0.0709} & \underline{0.6946 ± 0.1054} & \underline{0.6918 ± 0.0756} & \underline{0.7061 ± 0.1481} & \underline{0.6887 ± 0.0577} \\
\rowcolor{green!17.5}
KAN-KAN & \textbf{0.7091 ± 0.0464} & \textbf{0.6769 ± 0.1095} & \textbf{0.7108 ± 0.1075} & 0.6715 ± 0.0684 & \textbf{0.7688 ± 0.1637} & \textbf{0.6956 ± 0.0626} \\
\midrule
\rowcolor{gray!25}
\multicolumn{7}{c}{\textbf{\large FastKAN}} \\
\midrule
Configuration & ACC & AUC & F1 & PRE & SEN & SPE \\
\midrule
MLP-KAN & 0.6909 ± 0.0182 & 0.6274 ± 0.0966 & 0.7016 ± 0.0970 & 0.6497 ± 0.0397 & \underline{0.7927 ± 0.1909} & 0.6637 ± 0.0314 \\
\rowcolor{green!17.5}
KAN-MLP & \underline{0.7000 ± 0.0464} & \textbf{0.6563 ± 0.0927} & \textbf{0.7168 ± 0.0757} & \underline{0.6621 ± 0.0590} & \textbf{0.7938 ± 0.1284} & \textbf{0.6913 ± 0.0551} \\
KAN-KAN & \textbf{0.7091 ± 0.0464} & \underline{0.6471 ± 0.0287} & \underline{0.7123 ± 0.0926} & \textbf{0.6683 ± 0.0482} & 0.7745 ± 0.1641 & \underline{0.6845 ± 0.0381} \\
\midrule
\rowcolor{gray!25}
\multicolumn{7}{c}{\textbf{\large FasterKAN}} \\
\midrule
Configuration & ACC & AUC & F1 & PRE & SEN & SPE \\
\midrule
MLP-KAN & 0.6818 ± 0.0575 & 0.5815 ± 0.0838 & \underline{0.6955 ± 0.0824} & 0.6523 ± 0.0809 & \underline{0.7619 ± 0.1326} & 0.6635 ± 0.0424 \\
KAN-MLP & \underline{0.6909 ± 0.0340} & \textbf{0.6352 ± 0.0490} & 0.6716 ± 0.1045 & \underline{0.6776 ± 0.0630} & 0.6852 ± 0.1791 & \underline{0.6671 ± 0.0426} \\
\rowcolor{green!17.5}
KAN-KAN & \textbf{0.7091 ± 0.0464} & \underline{0.6203 ± 0.0490} & \textbf{0.7093 ± 0.1027} & \textbf{0.6813 ± 0.0761} & \textbf{0.7859 ± 0.2114} & \textbf{0.6777 ± 0.0278} \\
\midrule
\rowcolor{gray!25}
\multicolumn{7}{c}{\textbf{\large Wav-KAN}} \\
\midrule
Configuration & ACC & AUC & F1 & PRE & SEN & SPE \\
\midrule
\rowcolor{green!17.5}
MLP-KAN & \textbf{0.7727 ± 0.0287} & \textbf{0.7184 ± 0.0620} & \textbf{0.7682 ± 0.0482} & \underline{0.7840 ± 0.1086} & \textbf{0.7837 ± 0.1314} & \textbf{0.7768 ± 0.0270} \\
KAN-MLP & \underline{0.7545 ± 0.0464} & 0.6876 ± 0.0803 & 0.7204 ± 0.1090 & \textbf{0.8133 ± 0.1012} & \underline{0.7086 ± 0.2241} & \underline{0.7312 ± 0.0448} \\
KAN-KAN & 0.7364 ± 0.0445 & \underline{0.6921 ± 0.0547} & \underline{0.7218 ± 0.0457} & 0.7683 ± 0.0779 & 0.7032 ± 0.1246 & 0.7295 ± 0.0350 \\
\midrule
\rowcolor{gray!25}
\multicolumn{7}{c}{\textbf{\large ChebyKAN}} \\
\midrule
Configuration & ACC & AUC & F1 & PRE & SEN & SPE \\
\midrule
MLP-KAN & \underline{0.7091 ± 0.0464} & \textbf{0.6429 ± 0.0722} & \underline{0.7007 ± 0.1118} & 0.6766 ± 0.0527 & \underline{0.7409 ± 0.1825} & \underline{0.6859 ± 0.0528} \\
\rowcolor{green!17.5}
KAN-MLP & \textbf{0.7273 ± 0.0498} & \underline{0.6345 ± 0.1052} & \textbf{0.7443 ± 0.0838} & \underline{0.6801 ± 0.0374} & \textbf{0.8412 ± 0.1742} & \textbf{0.7061 ± 0.0794} \\
KAN-KAN & 0.6182 ± 0.0464 & 0.5741 ± 0.0487 & 0.4880 ± 0.2272 & \textbf{0.7278 ± 0.1458} & 0.5099 ± 0.3765 & 0.5604 ± 0.0416 \\
\bottomrule
\end{tabular}
}
\end{table*}

On the UM site, the ordering of the KAN variants by best average performance changes to: Wav-KAN $\rightarrow$ ChebyKAN $\rightarrow$ Efficient-KAN $\rightarrow$ FasterKAN $\rightarrow$ FastKAN. Wav-KAN in an MLP-KAN configuration achieves the best overall performance across all KAN variants and configurations. However, its comparatively poor performance under the same configuration on the NYU site indicates that the efficacy of KAN in replacing MLPs is heavily dependent on data characteristics. Also on the UM, there continues to be no consistent winner among the configurations, with Efficient-KAN and FasterKAN performing the best with a KAN-KAN configuration, FastKAN and ChebyKAN performing best in the KAN-MLP setting, and Wav-KAN performing the best under the MLP-KAN setting.

As a final experiment, we performed two experiments to examine how a hybrid configuration of FastKAN and Wav-KAN would perform across both the NYU and UM sites. The results of these experiments are shown in Table~\ref{tab:hybridresults}. In this table, we are only showing hybrid variations of the KAN-KAN configuration; we are not attempting to combine the functionality of both KAN methods into one.

\begin{table*}[t]
\centering
\caption{Exploration of hybrid KAN configurations within the ABFR-KAN framework. Specifically, FastKAN and Wav-KAN hybrids. Mean$\pm$stdev scores are reported by averaging the results of 5-fold cross-validation. Best and second-best results are \textbf{bolded} and \underline{underlined}, respectively.}
\label{tab:hybridresults}
\resizebox{\linewidth}{!}{
\begin{tabular}{lcccccc}
\toprule
\rowcolor{gray!25}
\multicolumn{7}{c}{\textbf{\large NYU Site Classification Performance}} \\
\midrule
Configuration & ACC & AUC & F1 & PRE & SEN & SPE \\
\midrule
\rowcolor{green!17.5}
FastKAN-FastKAN & \textbf{0.7427 ± 0.0342} & \textbf{0.7041 ± 0.0708} & \textbf{0.8003 ±  0.0200} & 0.7267 ± 0.0469 & \textbf{0.8984 ± 0.0646} & \textbf{0.7159 ± 0.0434} \\
FastKAN-WavKAN & 0.6897 ± 0.0324 & 0.6327  ± 0.1084 & \underline{0.7472 ± 0.0379} & 0.7079 ± 0.0555 & \underline{0.8147 ± 0.1335} & 0.6664 ± 0.0522 \\
WavKAN-FastKAN & \underline{0.7017 ± 0.0394} & \underline{0.6797 ± 0.0606} & 0.7444 ± 0.0408 & \underline{0.7327 ± 0.0445} & 0.7658 ± 0.0921 & \underline{0.6924 ± 0.0388} \\
WavKAN-WavKAN & 0.6726 ± 0.0201 & 0.6466 ± 0.0185 & 0.6933 ± 0.0542 & \textbf{0.7540 ± 0.0750} & 0.6637 ± 0.1157 & 0.6733 ± 0.0181 \\
\midrule
\rowcolor{gray!25}
\multicolumn{7}{c}{\textbf{\large UM Site Classification Performance}} \\
\midrule
Configuration & ACC & AUC & F1 & PRE & SEN & SPE \\
\midrule
FastKAN-FastKAN & \underline{0.7091 ± 0.0464} & 0.6471 ± 0.0287 & 0.7123 ± 0.0926 & \underline{0.6683 ± 0.0482} & 0.7745 ± 0.1641 & 0.6845 ± 0.0381 \\
FastKAN-WavKAN & 0.7000 ± 0.0617 & 0.6158 ± 0.1216 & \textbf{0.7220 ± 0.0934} & 0.6569 ± 0.0821 & \textbf{0.8291 ± 0.1836} & 0.6734 ± 0.0615 \\
WavKAN-FastKAN & 0.6909 ± 0.0340 & \textbf{0.7059 ± 0.0835} & 0.7177 ± 0.0439 & 0.6613 ± 0.0731 & \underline{0.8103 ± 0.1288} & \underline{0.6979 ± 0.0370} \\
\rowcolor{green!17.5}
WavKAN-WavKAN & \textbf{0.7364 ± 0.0445} & \underline{0.6921 ± 0.0547} & \underline{0.7218 ± 0.0457} & \textbf{0.7683 ± 0.0779} & 0.7032 ± 0.1246 & \textbf{0.7295 ± 0.0350} \\
\bottomrule
\end{tabular}
}
\end{table*}

On both ABIDE I sites, the hybrid configurations do not manage to outperform the respective best-performing single KAN variants. On the NYU site, FastKAN-WavKAN and WavKAN-FastKAN both fall short of outperforming FastKAN-FastKAN, although both methods end up outperforming the baseline WavKAN-WavKAN setting, with WavKAN-FastKAN (so, WavKAN in the ViT encoder, FastKAN as the classification head) performing better than its inverse version. On the UM site, WavKAN-WavKAN offers the best performance, with the hybrid versions failing to surpass it. But again, the hybrid configurations do show some promise in outperforming the other baseline configuration, FastKAN-FastKAN, with FastKAN-WavKAN performing slightly better than WavKAN-FastKAN. From these results, two things are clear. \textit{First,} the hybrid configurations are not suitable for outperforming the best-performing single KAN-KAN configurations; and \textit{second}, they do show some promise in boosting the performance of less optimally performing methods. This indicates that even though we deem hybrid KAN configurations not suitable for further pursuit for our specific task of ASD vs. NC from rs-fMRI data, there may still be use cases for them in other tasks, and this warrants further exploration.

\end{document}